\documentclass[10pt,twocolumn,letterpaper,table]{article}

\usepackage[]{cvpr} 
\usepackage[dvipsnames]{xcolor}

\usepackage{booktabs, multirow} % for borders and merged ranges
\usepackage{soul}% for underlines

\newcommand{\eclipse}{\textit{ECLIPSE}}

\DeclareMathOperator{\E}{\mathbb{E}}

\definecolor{bluegray}{rgb}{0.67, 0.9, 0.93}
\definecolor{babypink}{rgb}{0.96, 0.76, 0.76}
\definecolor{lightgreen}{rgb}{0.56, 0.93, 0.56}
\definecolor{mygreen}{rgb}{0.0, 0.5, 0.0}

\newcommand{\nresults}[1]{\textsubscript{{\color{BrickRed}\textbf{#1}}}}

\definecolor{cvprblue}{rgb}{0.21,0.49,0.74}
\usepackage[pagebackref,breaklinks,colorlinks,citecolor=cvprblue]{hyperref}

\title{\eclipse: A Resource-Efficient Text-to-Image Prior for Image Generations}

\author{Maitreya Patel,~~Changhoon Kim,~~Sheng Cheng,~~Chitta Baral,~~Yezhou Yang\\
Arizona State University\\
{\tt\small \{maitreya.patel, kch, scheng53, chitta, yz.yang\}@asu.edu}
}

\begin{document}

\maketitle

\begin{abstract}

Text-to-image (T2I) diffusion models, notably the unCLIP models (e.g., DALL-E-2), achieve state-of-the-art (SOTA) performance on various compositional T2I benchmarks, at the cost of significant computational resources. 
The unCLIP stack comprises T2I prior and diffusion image decoder. 
The T2I prior model alone adds a billion parameters compared to the Latent Diffusion Models, which increases the computational and high-quality data requirements. 
We introduce \eclipse\footnote{Our strategy, \eclipse, draws an analogy from the way a smaller prior model, akin to a celestial entity, offers a glimpse of the grandeur within the larger pre-trained vision-language model, mirroring how an eclipse reveals the vastness of the cosmos.}, a novel contrastive learning method that is both parameter and data-efficient. 
\eclipse~leverages pre-trained vision-language models (e.g., CLIP) to distill the knowledge into the prior model.
We demonstrate that the \eclipse~trained prior, with only 3.3\% of the parameters and trained on a mere 2.8\% of the data, surpasses the baseline T2I priors with an average of 71.6\% preference score under resource-limited setting. 
It also attains performance on par with SOTA big models, achieving an average of 63.36\% preference score in terms of the ability to follow the text compositions.
Extensive experiments on two unCLIP diffusion image decoders, Karlo and Kandinsky,
affirm that \eclipse~priors consistently deliver high performance while significantly reducing resource dependency.
Project page: \href{https://eclipse-t2i.vercel.app/}{https://eclipse-t2i.vercel.app/}

\end{abstract}

\section{Introduction}
\label{sec:intro}

Diffusion models~\cite{sohl2015deep,ho2020denoising, ramesh2022hierarchical, rombach2022high} have demonstrated remarkable success in generating high-quality images conditioned on text prompts.
This Text-to-Image (T2I) generation paradigm has been effectively applied to various downstream tasks such as subject/segmentation/depth-driven image generation~\cite{chefer2023attend, chen2023training, patel2023conceptbed, gal2022image, li2023gligen}. 
Central to these advancements are two predominant text-conditioned diffusion models: Latent Diffusion Models (LDM)~\cite{rombach2022high}, also known as Stable Diffusion, and unCLIP models~\cite{ramesh2022hierarchical}.
The LDM, notable for its open-source availability, has gained widespread popularity within the research community.
On the other hand, unCLIP models have remained under-studied.
Both model types fundamentally focus on training the diffusion models conditioned on text prompts.
The LDM contains a singular text-to-image diffusion model, while unCLIP models have a text-to-image prior, and a diffusion image decoder.
Both model families work within the vector quantized latent space of the image~\cite{van2017neural}.
In this paper, we focus on unCLIP models because they consistently outperform other SOTA models in various composition benchmarks such as T2I-CompBench~\cite{huang2023t2i} and HRS-Benchmark~\cite{bakr2023hrs}.

\begin{figure}
  \includegraphics[width=\linewidth]{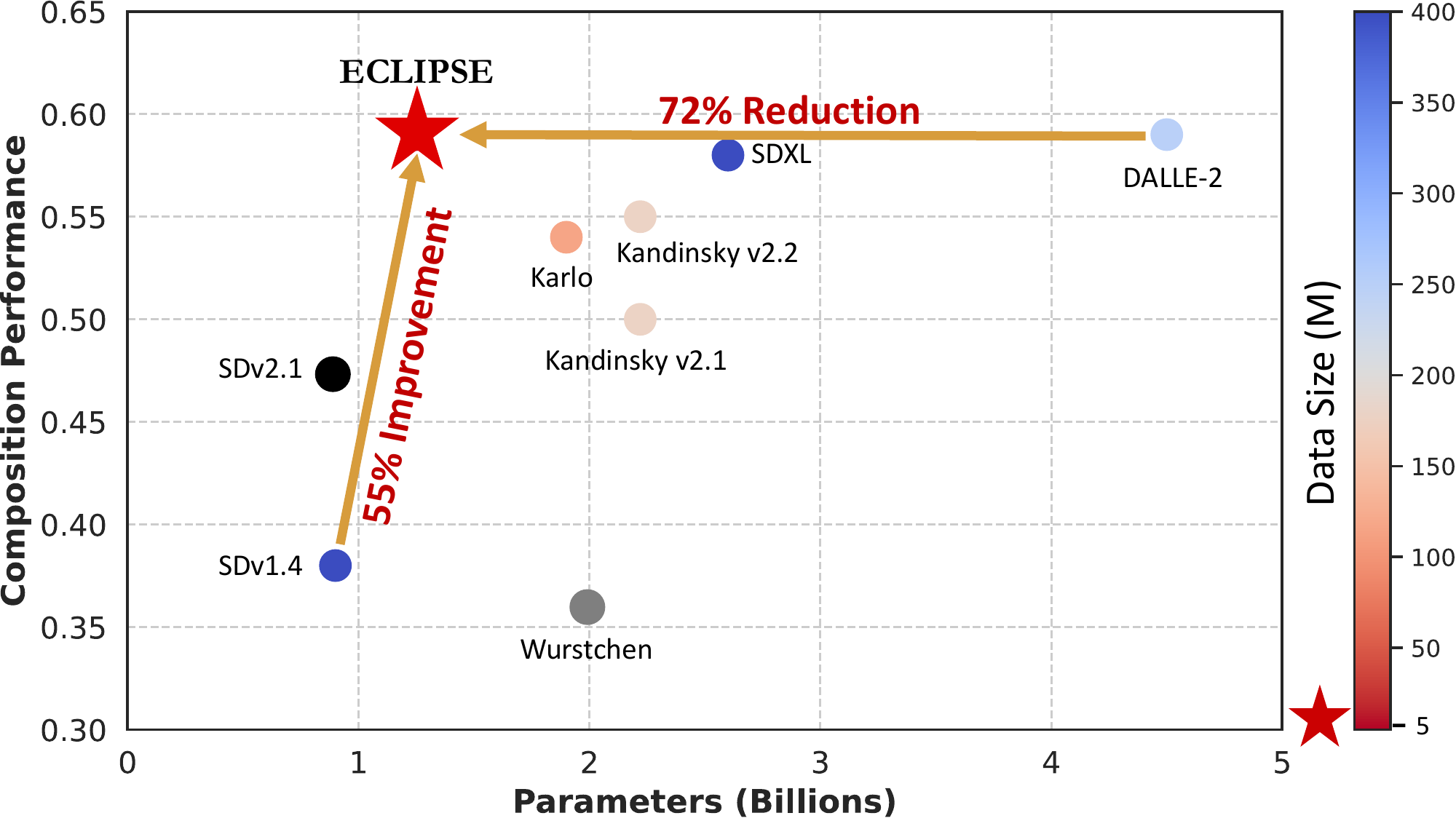}
  \caption{
  Comparison between SOTA text-to-image models with respect to their total number of parameters and the average performance on the three composition tasks (color, shape, and texture).
  \eclipse~achieves better results with less number of parameters without requiring a large amount of training data. The shown \eclipse~trains a T2I prior model (having only 33M parameters) using only 5M image-text pairs with Kandinsky decoder.
  }
  \label{fig:teaser_results}
\end{figure}

These T2I models, typically large in parameter count, require massive amounts of high-quality image-text pairs for training.
unCLIP models like DALL-E-2~\cite{ramesh2022hierarchical}, Karlo~\cite{kakaobrain2022karlo-v1-alpha}, and Kandinsky~\cite{razzhigaev2023kandinsky}, feature prior module containing approximately 1 billion parameters, resulting in a significant increase in overall model size ($\geq$ 2B) compared to LDMs.
These unCLIP models are trained on 250M, 115M, and 177M image-text pairs, respectively.
Therefore, two critical questions remain: 
\textit{{ 1) Does the incorporation of a text-to-image prior contribute to SOTA performance on text compositions? 
2) Or is scaling up model size the key factor?} }
In this study, we aim to deepen the understanding of T2I priors and propose substantial enhancements to existing formulations by improving parameter and data efficiency.

As proposed by~\citet{ramesh2022hierarchical}, T2I priors are also diffusion models, which are designed to directly estimate the noiseless image embedding at any timestep of the diffusion process.
We perform an empirical study to analyze this prior diffusion process.
We find that this diffusion process has a negligible impact on generating accurate images and having the diffusion process slightly hurts the performance.
Moreover, diffusion models require substantial GPU hours/days for training due to the slower convergence.
Therefore, in this work, we use the non-diffusion model as an alternative. 
While this approach may reduce the compositional capabilities due to the absence of classifier-free guidance~\cite{ho2022classifier}, it significantly enhances parameter efficiency and decreases the dependencies on the data.

To overcome the above limitations, in this work, we introduce \eclipse, a novel contrastive learning strategy to improve the T2I non-diffusion prior. 
We improve upon the traditional method of maximizing the Evidence Lower Bound (ELBO) for generating the image embedding from the given text embedding.
We propose to utilize the semantic alignment (between the text and image) property of the pre-trained vision-language models to supervise the prior training.
Utilizing \eclipse, we train compact (97\% smaller) non-diffusion prior models (having 33 million parameters) using a very small portion of the image-text pairs (0.34\% - 8.69\%).
We train \eclipse~priors for two unCLIP diffusion image decoder variants (Karlo and Kandinsky).
The \eclipse-trained priors significantly surpass baseline prior learning strategies and rival the performance of 1 billion parameter counterparts.
Our results indicate a promising direction for T2I generative models, achieving better compositionality without relying on extensive parameters or data.
As illustrated in Fig.~\ref{fig:teaser_results}, by simply improving the T2I prior for unCLIP families, their overall parameter and data requirements drastically reduce and achieve the SOTA performance against similar parameter models.

\medskip\noindent\textbf{Contributions.}
1) We introduce \eclipse, the first attempt to employ contrastive learning for text-to-image priors in the unCLIP framework. 
2) Through extensive experimentation, we demonstrate \eclipse's superiority over baseline priors in resource-constrained environments.
3) Remarkably, \eclipse~priors achieve comparable performance to larger models using only 2.8\% of the training data and 3.3\% of the model parameters.
4) We also analyze and offer empirical insights on the shortcomings of existing T2I diffusion priors.

\section{Related Works}
\label{sec:formatting}

\noindent\textbf{Text-to-Image Generative Models.}
Advancements in vector quantization and diffusion modeling have notably enhanced text-to-image generation capabilities. 
Notable works like DALL-E~\cite{ramesh2021zero} have leveraged transformer models trained on quantized latent spaces. 
Contemporary state-of-the-art models, including GLIDE~\cite{nichol2021glide}, Latent Diffusion Model (LDM)~\cite{rombach2022high}, DALL-E-2~\cite{ramesh2022hierarchical}, and Imagen~\cite{saharia2022photorealistic}, have significantly improved over earlier approaches like StackGAN~\cite{zhang2017stackgan} and TReCS~\cite{koh2021text}. 
As these models achieve remarkable photorealism, several works focus on making T2I models more secure~\cite{kim2023wouaf, stable_signature, nie2023attributing, kim2020decentralized}.
LDM models primarily focus on unified text-to-image diffusion models that incorporate the cross-attention layers~\cite{rombach2022high}. 
Additionally, several studies aim at refining Stable Diffusion models during inference through targeted post-processing strategies~\cite{chefer2023attend, chen2023training, phung2023grounded}. 
In contrast, unCLIP models, exemplified by DALL-E-2~\cite{kim2023architectural}, Karlo~\cite{kakaobrain2022karlo-v1-alpha}, and Kandinsky~\cite{razzhigaev2023kandinsky}, incorporate a two-step process of text-to-image diffusion transformer prior model and diffusion image decoder having the same model architecture as LDMs. 
Recent benchmarks have highlighted the superior compositional capabilities of DALL-E-2 over LDM methods~\cite{huang2023t2i,bakr2023hrs}. 
Our work examines and enhances existing prior learning strategies in open-source pre-trained unCLIP models, Karlo and Kandinsky.

\begin{figure*}[!t]
  \includegraphics[width=\textwidth]{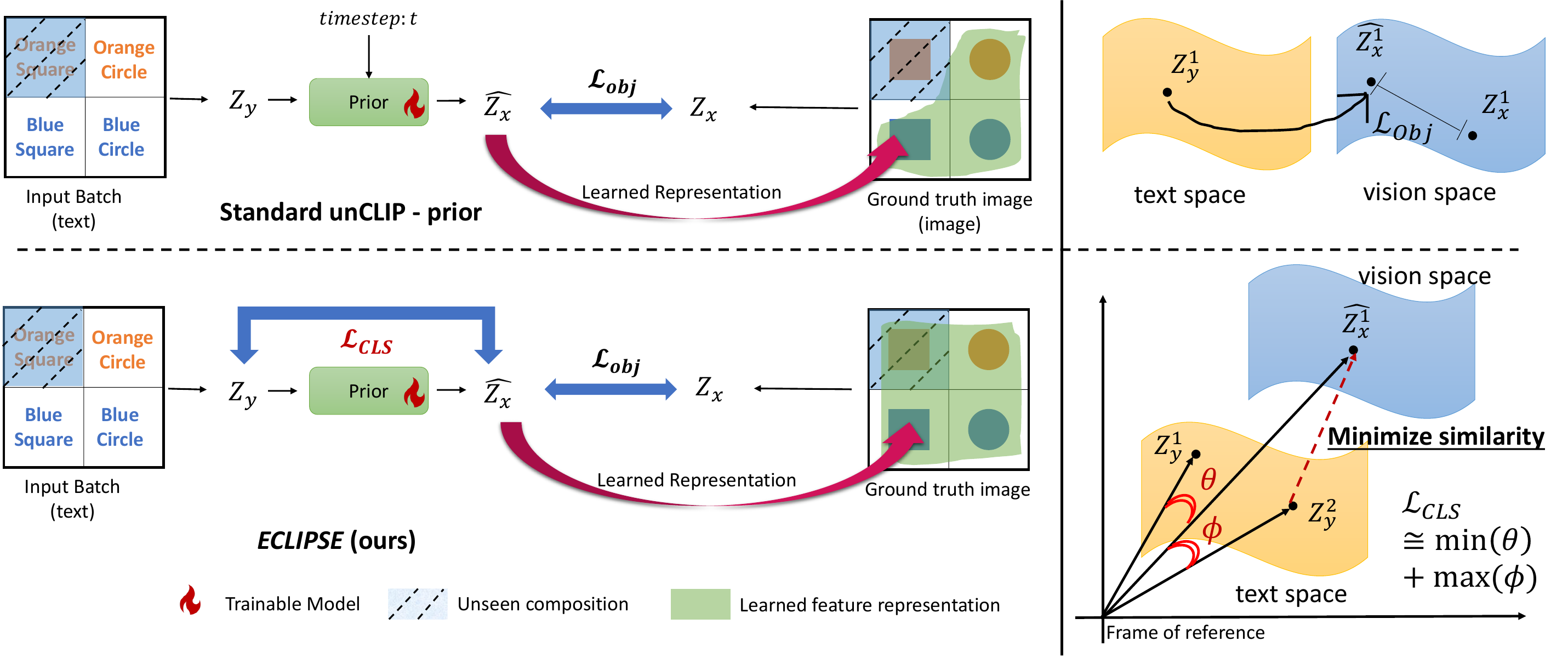}
  \caption{Standard T2I prior learning strategies (top) minimizes the mean squared error between the predicted vision embedding $\hat{z}_x$ w.r.t. the ground truth embedding $z_x$ with or without time-conditioning. This methodology cannot be generalized very well to the outside training distribution (such as Orange Square). The proposed \eclipse~training methodology (bottom) utilizes the semantic alignment property between $z_x$ and $z_y$ with the use of contrastive learning, which improves the text-to-image prior generalization.}
  \label{fig:main_method}
\end{figure*}

\medskip\noindent\textbf{Efficient Text-to-Image Models.}
The current generation of T2I models is characterized by extensive parameter sizes and demanding training requirements, often necessitating thousands of GPU days. 
Research efforts have primarily centered on model refinement through knowledge distillation, step distillation, and architectural optimization~\cite{li2023snapfusion, salimans2021progressive, luo2023latent}. 
Wuerstchen~\cite{pernias2023wuerstchen} presents an efficient unCLIP stack requiring fewer training time GPU hours. 
Concurrently, Pixart-$\alpha$~\cite{chen2023pixart} leverages pre-trained Diffusion-Transformers (DiT)~\cite{peebles2023scalable} as base diffusion models, further reducing training time. 
Distinctively, \eclipse~focuses on refining text-to-image priors within the unCLIP framework using a mere 3.3\% of the original model parameters, thereby significantly reducing the training duration to approximately 200 GPU hours. 
Our work falls orthogonal to the existing efficient T2I methodologies that mainly focus on knowledge and step distillation, and/or architectural compression.
When integrated with these model compression strategies, \eclipse~can position the unCLIP family models as a compact yet highly accurate and efficient T2I generation methodology.

\medskip\noindent\textbf{Contrastive Learning in Generative Models.}
Contrastive learning, traditionally applied in visual discriminative tasks, has seen utilization in image-text alignment models like CLIP~\cite{radford2021learning}, LiT~\cite{zhai2022lit}, and SigLIP~\cite{zhai2023sigmoid}. 
However, its application in generative models, particularly in Generative Adversarial Networks (GANs), remains limited~\cite{zhou2022towards, li2022stylet2i, cong2023attribute}. For instance, Lafite~\cite{zhou2022towards} employs a contrastive approach for image-to-text prior training in language-free T2I GANs. 
StyleT2I~\cite{li2022stylet2i} attempts to learn the latent edit direction for StyleGAN~\cite{karras2019style}, which is supervised via spatial masks on the images making the method not scalable.
ACTIG~\cite{cong2023attribute} introduces an attribute-centric contrastive loss to enhance discriminator performance. 
These methods are constrained by their domain-specific knowledge requirements and inability to be directly applied to diffusion models~\cite{li2022stylet2i,cong2023attribute}. 
In contrast, \eclipse~applies CLIP-based contrastive learning to train more effective T2I prior models in diffusion-based T2I systems. 
This strategy is not only resource-efficient but significantly enhances the traditional text-to-image diffusion priors by exploiting the semantic latent space of pre-trained vision-language models.

\section{Methodology}
\label{sec:methodology}

This section elaborates on the Text-to-Image (T2I) methodologies, beginning with an overview of unCLIP, followed by the formal problem statement. 
We then delve into our proposed training strategy, \eclipse, for T2I prior in detail.
Figure~\ref{fig:main_method} provides the overview of baselines and \eclipse~training strategies.

\subsection{Preliminaries}
\label{sec:unclip}

Without the loss of generality, let's assume that $y \in Y$ denotes the raw text and $x \in X$ denotes the raw image.
$z_x$ and $z_y$ denote the image and text latent embeddings extracted using the pre-trained vision and text encoders ($z_x = C_{vision}(x); \quad z_y = C_{text}(y)$).
Ideally, these $C_{text}$ and $C_{vision}$ can be any model (e.g., T5-XXL, ViT, and CLIP).
Both model families (LDM and unCLIP) fundamentally focus on learning a mapping function $f_{\theta}: Y \rightarrow X$.
The LDMs contain a singular text-to-image decoder model ($f_\theta$), while unCLIP framework ($f_\theta = h_\theta \circ g_\phi$) contains two primary modules:

\begin{itemize}
    \item \textbf{Text-to-Image Prior} ($g_\phi : z_y \rightarrow z_x$): 
    This module maps the text embeddings to the corresponding vision embeddings.
    \citet{ramesh2022hierarchical} showed that the diffusion model as T2I prior leads to slightly better performance than the autoregressive models. 
    For each timestep $t$ and a noised image embedding $z_x^{(t)}\sim q(t, z_x)$ (here, $q$ is a forward diffusion process), the diffusion prior directly estimates noiseless $z_x$ rather than estimating Gaussian noise distribution $\epsilon \sim \mathcal{N}(0,\mathcal{I})$ as: 
\end{itemize}
\begin{equation}
    \mathcal{L}_{prior} = 
    \underset{
        \substack{
            t \sim [0, T], \\
            z_x^{(t)} \sim q(t, z_x)
            }
        }{\E}
        \Big[||z_x - g_{\phi}(z_x^{(t)},t,z_y)||_{2}^{2} \Big].
    \label{eq:prior}
\end{equation}

\begin{itemize}
    \item \textbf{Diffusion Image Decoder} ($h_\theta : (z_x, z_y) \rightarrow x$): 
    This module generates the final image conditioned on the $z_x$ and the input text features $z_y$. 
    This diffusion decoder follows the standard diffusion training procedure by estimating $\epsilon \sim \mathcal{N}(0, \mathcal{I})$ after~\cite{ho2020denoising}:
\end{itemize}
\begin{equation}
    \mathcal{L}_{decoder} = 
    \underset{
        \substack{
            \epsilon \sim N(0, I) \\
            t \sim [0, T], \\
            (z_x,~z_y)
            }
        }{\E}
        \Big[||\epsilon - h_{\theta}(x^{(t)},t,z_x,z_y)||_{2}^{2} \Big].
    \label{eq:decoder}
\end{equation}

Specifically, different versions of the unCLIP decoder, such as Kandinsky and Karlo, vary in whether they include text conditioning ($z_y$) in the diffusion image decoder. 
However, both approaches yield comparable results, provided that image conditioning ($z_x$) is accurate.
Both training objectives, $\mathcal{L}_{prior}$ and $\mathcal{L}_{decoder}$, integrate Classifier-Free Guidance (CFG)~\cite{ho2022classifier}, enhancing the model's generative capabilities.
During training, conditions are omitted 10\% of the time to foster unconditional generation, subsequently improving test performance as CFG works as implicit classifier guidance~\cite{ho2022classifier}.

\subsection{Problem Formulation}

Given the pivotal role of the T2I prior module in image generation from text, in this paper, our focus is on enhancing $g_\phi$, while keeping the pre-trained $h_\theta$ frozen.
Let's consider a training distribution $P_{XY}$, comprising input pairs of image and text $(x, y)$.
Maximizing the Evidence Lower Bound (ELBO) on the training distribution $P_{XY}$ facilitates this mapping of $z_y \rightarrow z_x$.
However, such a strategy does not inherently assure generalization, especially when the input text prompt ($y$) deviates from the assumed independently and identically distributed (i.i.d.) pattern of $P_{XY}$~\cite{vapnik1991principles}.
Therefore, attaining a more diverse and representative $P_{XY}$ becomes crucial for improving the performance.
While a diffusion prior combined with CFG has been shown to bolster generalization, especially with diverse training data and extensive training iterations~\cite{okawa2023compositional}, it is computationally expensive and is not always reliable (especially, in low resource constraint settings) as shown in Section~\ref{sec:quantitative_evals}.
Given these constraints, our goal is to develop an alternative prior learning methodology that not only improves parameter efficiency (97\% reduction) and mitigates the need for large-scale high-quality data ($\leq 5\%$) but also upholds performance levels.

\subsection{Proposed Method: \textbf{\eclipse}}

This section elaborates on \eclipse, our model training strategy to learn text-to-image prior ($g_\phi$).
We focus on enhancing non-diffusion prior models through the effective distillation of pre-trained vision-language models, such as CLIP, while preserving the semantic alignment between the input text embedding $z_y$ and corresponding estimated vision embeddings $\hat{z}_x$ by using the contrastive loss.

\medskip\noindent\textbf{Base Prior Model.}
T2I diffusion prior deviates from the standard diffusion objective (such as Eq.~\ref{eq:decoder}).
Unlike the standard $\epsilon \sim \mathcal{N}(0, \mathcal{I})$ prediction-based diffusion objective, the T2I diffusion prior objective (Eq.~\ref{eq:prior}) do not compare two Gaussian distributions, instead, it directly estimates the $z_x$ which is noiseless.
However, during inference, we still adhere to the conventional denoising process, introducing additional noise ($\sigma_t\epsilon$) at each step, except for the final step according to \citet{ho2020denoising}.
This creates a new input distribution ($z_x + \sigma_t\epsilon$), possibly unencountered during training.
Moreover, if we repeat this for $T$ timesteps, it can lead to the accumulation of errors, which is undesirable. 
We provide empirical analysis in Section~\ref{sec:ablation} to ground this hypothesis, where we show that having more diffusion prior steps does not benefit the overall text-to-image generation abilities. 

Therefore, to mitigate this unnecessary computing, we use non-diffusion T2I prior, making the prior model both parameter-efficient and less demanding in terms of computational resources. 
This non-diffusion architecture forms our base model, and we introduce the training objective that leverages pre-trained vision-language models trained on extensive datasets to improve generalization outside the $P_{XY}$ distribution.

\medskip\noindent\textbf{Projection Objective.}
Despite vision-language models aligning the semantic distributions across modalities, each modality may exhibit unique distributions.
Therefore, our approach involves projecting the text embedding onto the vision embedding.
This is achieved using a mean squared error objective between the predicted vision embedding ($\hat{z}_x$) and the ground truth vision embedding ($z_x$):
\begin{equation}
\label{eq:projection}
    \mathcal{L}_{proj} = 
    \underset{
        \substack{
            \epsilon \sim \mathcal{N}(0,I) \\
            z_y, z_x
        }
    }{\E} \Big[ || z_x - g_\phi(\epsilon, z_y)||_2^2 \Big],
\end{equation}

\noindent where $\epsilon$ is the Gaussian input noise.
Notably, as discussed previously, this is an approximation of the diffusion prior objective (Eq.~\ref{eq:prior}) with $t=T$ and without CFG.
$\mathcal{L}_{proj}$ learns latent posterior distribution with the \textit{i.i.d.} data assumption.
However, this model, fine-tuned on $P_{XY}$, may not generalize well beyond its distribution.
The optimal solution would be to train on a dataset that encapsulates all potential distributions to cover all possible scenarios, which is an impractical and resource-consuming task.

\paragraph{CLIP Contrastive Learning.}

\begin{table*}[!htb]
\centering
\caption{
The comparison (in terms of FID and compositions) of the baselines and state-of-the-art methods with respect to the \eclipse. 
* indicates the official reported ZS-FID.
$\Psi$ denotes the FID performance of a model trained on MSCOCO.
The best performing \eclipse~variant (with respect to its big counterpart) is highlighted by 
\colorbox{ForestGreen!40}{green}.
% {\textcolor{ForestGreen}{green}}.
\eclipse~consistently outperforms the SOTA big models despite being trained on a smaller subset of dataset and parameters.
% \colorbox{bluegray}{Blue} denotes the best performing model utilizing the Karlo diffusion image decoder.
% Similarly, \colorbox{babypink}{red} denotes the best model using Kandinsky v2.2 decoder.
% Model parameters and data size is in millions.
% Evaluations are on MSCOCO zero-shot FID, and various compositionality categories from T2I Benchmark~\cite{huang2023t2i}.
% The first two blocks shows the comparison with respect to the Karlo and Kandinsky v2.1 image decoders.
% While the third block contains the other standard baselines for overall comparison. 
% We highlight the overall state-of-the-art with \colorbox{F4C2C2}{red} color. 
% We utilize \colorbox{bluegray}{\textbf{blue}} to highlight the block specific best performing methods.
% Our proposed method is denoted as \colorbox{lightgreen}{\textbf{\eclipse}}.
}

\label{tab:main_results}

\scriptsize

\resizebox{\textwidth}{!}{%

\begingroup
\setlength{\tabcolsep}{4pt}

\begin{tabular}{l | c | ccc | c | cccccc}\toprule

\multirow{2}{*}{\textbf{Methods}} & \textbf{Model} &\textbf{Training} &\textbf{Total} &\textbf{Data} &\textbf{ZS-} &\multicolumn{5}{c}{\textbf{T2I-CompBench}} \\

% \cmidrule{7-11}

 & \textbf{Type}  &\textbf{Params [M]*} &\textbf{Params [B]} &\textbf{Size [M]} &\textbf{FID ($\downarrow$)} &\textbf{Color ($\uparrow$)} &\textbf{Shape ($\uparrow$)} &\textbf{Texture ($\uparrow$)} &\textbf{Spatial ($\uparrow$)} &\textbf{Non-Spatial ($\uparrow$)} \\\midrule

% GALIP &GAN &330 &0.33 &12 & 12.54\textsuperscript{*} & & & & & \\
Stable Diffusion v1.4 &LDM &900 &0.9 &400 & 16.31\textsuperscript{*} &0.3765 &0.3576 &0.4156 &0.1246 &0.3076 \\
Stable Diffusion v2.1 &LDM &900 &0.9 &2000 & 14.51\textsuperscript{*} &0.5065 &0.4221 &0.4922 &0.1342 &0.3096 \\

Wurstchen & unCLIP & 1000 & 2.0 & 1420 & 23.60\textsuperscript{*} & 0.3216 & 0.3821 & 0.3889 & 0.0696 & 0.2949 \\

Kandinsky v2.1 &unCLIP &1000 &2.22 &177 & 18.09~ &0.4647 &0.4725 &0.5613 & 0.1219 & 0.3117 \\

DALL-E-2 &unCLIP &1000 &4.5 &250 &10.65\textsuperscript{*} &0.5750 &0.5464 &0.6374 &0.1283 &0.3043
\\
\midrule

Karlo &unCLIP &1000 &1.9 &115 &\cellcolor{ForestGreen!50}20.64~ &0.5127 &\cellcolor{ForestGreen!50}0.5277 & 0.5887 &0.1337 &0.3112 \\

\cellcolor{Apricot!50} & \multirow{3}{*}{Karlo} &\textbf{33} &\textbf{0.93} &\textbf{0.6}\nresults{MSCOCO} & 23.67\textsuperscript{$\Psi$} &\cellcolor{ForestGreen!50}0.5965 &0.5063 &\cellcolor{ForestGreen!50}0.6136 & \cellcolor{ForestGreen!50}0.1574 & \cellcolor{ForestGreen!50}0.3235 \\
\cellcolor{Apricot!50}& &\textbf{33} &\textbf{0.93} &\textbf{2.5}\nresults{CC3M} &  26.73 &\cellcolor{ForestGreen!50}0.5421 &0.5090 &0.5881 & \cellcolor{ForestGreen!50}0.1478 & \cellcolor{ForestGreen!50}0.3213  \\
\cellcolor{Apricot!50}\multirow{-3}{*}{\textbf{\eclipse~(ours)}}& &\textbf{33} &\textbf{0.93} &\textbf{10.0}\nresults{CC12M} &26.98 &\cellcolor{ForestGreen!50}0.5660 &0.5234 &\cellcolor{ForestGreen!50}0.5941 &\cellcolor{ForestGreen!50}0.1625 &\cellcolor{ForestGreen!50}0.3196 \\

\midrule
Kandinsky v2.2 &unCLIP &1000 &2.22 &177 & 20.48~ &0.5768 & 0.4999 &0.5760 &\cellcolor{ForestGreen!50}0.1912 &0.3132 \\

\cellcolor{Apricot!50} & \multirow{2}{*}{Kandinsky v2.2} & \textbf{34} &\textbf{1.26} &\textbf{0.6}\nresults{MSCOCO} & 16.53\textsuperscript{$\Psi$} & \cellcolor{ForestGreen!50}0.5785 & 0.4951 & \cellcolor{ForestGreen!50}0.6173 & 0.1794 & \cellcolor{ForestGreen!50}0.3204 \\
\cellcolor{Apricot!50}\multirow{-2}{*}{\textbf{\eclipse~(ours)}} &  & \textbf{34} &\textbf{1.26} &\textbf{5.0}\nresults{HighRes} & \cellcolor{ForestGreen!50}19.16 & \cellcolor{ForestGreen!50}0.6119 & \cellcolor{ForestGreen!50}0.5429 & \cellcolor{ForestGreen!50}0.6165 & 0.1903 & \cellcolor{ForestGreen!50}0.3139\\

\bottomrule

\end{tabular}

\endgroup

}

\end{table*}

To address these limitations, we propose utilizing the CLIP more effectively, which contains the semantic alignment between image and language.
Specifically, we apply the CLIP Contrastive Loss after~\cite{radford2021learning} to train the T2I priors. 
For a given input batch $\{(z_x^i, z_y^i)\}_{i=1}^N$ from the $P_{XY}$ distribution, we calculate the text-conditioned image contrastive loss for the $i^{th}$ image embedding prediction relative to the all input ground truth text embeddings as:

\begin{equation}
\label{eq:clip}
    \mathcal{L}_{CLS;~ y \rightarrow x} = 
        -\frac{1}{N} \sum_{i=0}^{N} \log \frac{\exp(\langle \hat{z}^i_x, z^i_y \rangle/\tau)}{\sum_{j \in [N]} \exp(\langle \hat{z}^i_x, z^j_y\rangle/\tau)},
\end{equation}

\noindent where $\tau$ is the temperature parameter, $\langle , \rangle$ denotes the cosine similarity, and $N$ is the batch size.
This loss encourages the model to understand and follow the input text better, effectively reducing overfitting to the $P_{XY}$, as illustrated in Figure~\ref{fig:main_method}. 
Consequently, the final objective function is:
\begin{equation}
    \mathcal{L}_{ECLIPSE} = \mathcal{L}_{proj} + \lambda*\mathcal{L}_{CLS;~y \rightarrow x},
    \label{eq:objective}
\end{equation}

\noindent where $\lambda$ is the hyperparameter balancing the regularizer’s effect. 
Overall, the final objective function aims to map the text latent distribution to the image latent distribution via $\mathcal{L}_{proj}$ and such that it preserves the image-text alignment using $\mathcal{L}_{CLS;~y \rightarrow x}$.
This makes the prior model generalize beyond the given training distribution $P_{XY}$ such that it can follow the semantic alignment constraint. 
Importantly, we cannot use $\mathcal{L}_{CLS;~y \rightarrow x}$ alone or with a high value of $\lambda$ as the prior model will converge outside the vision latent distribution that optimizes the contrastive loss (such input text latent space itself).
And keeping $\lambda$ to a very low value cannot do knowledge distillation well enough.
Empirical studies suggest setting $\lambda=0.2$ for optimal performance, balancing knowledge distillation, and maintaining alignment within the vision latent distribution.

\section{Experiments \& Results}

This section introduces the datasets, training specifications, comparative baselines, and evaluation metrics utilized in our experiments. 
We conduct an extensive assessment of our proposed \eclipse~methodology and its variants, both quantitatively and qualitatively.

\subsection{Experimental Setup}

\medskip\noindent\textbf{Dataset.} Our experiments span four datasets of varying sizes: MSCOCO~\cite{lin2014microsoft}, CC3M~\cite{sharma2018conceptual}, CC12M~\cite{changpinyo2021conceptual}, and LAION-HighResolution\footnote{\url{https://huggingface.co/datasets/laion/laion-high-resolution}}~\cite{schuhmann2022laion}.
MSCOCO comprises approximately 0.6 million image-text pairs, while CC3M and CC12M contain around 2.5 and 10 million pairs, respectively
\footnote{According to the download date: 08/26/2023}
.
We select a very small subset of 5 million (2.8\%) image-text pairs from the LAION-HighRes dataset (175M).
We perform Karlo diffusion image decoder-related experiments on MSCOCO, CC3M, and CC12M as these datasets are subsets of the data used to train the Karlo diffusion image decoder.
Similarly, we use MSCOCO and LAION-HighRes for the Kandinsky decoder. 

\medskip\noindent\textbf{Baselines.}
\eclipse~variants are compared against leading T2I models, including Stable Diffusion, Wurstchen, Karlo, Kandinsky, and DALL-E-2. 
Additionally, we introduce two more baselines along with \eclipse~to evaluate the impact of our training strategy in a resource-constrained environment:
1) Projection: A non-diffusion prior model trained with $\mathcal{L}_{proj}$ (Eq.~\ref{eq:projection}).
2) Diffusion-Baseline: A diffusion prior model trained with $\mathcal{L}_{prior}$ (Eq.~\ref{eq:prior}) -- the traditional T2I prior, and
3) \eclipse: A non-diffusion prior model trained with our proposed methodology $\mathcal{L}_{ECLIPSE}$ (Eq.~\ref{eq:objective}).

\begin{figure*}[!ht]
    \includegraphics[width=\linewidth]{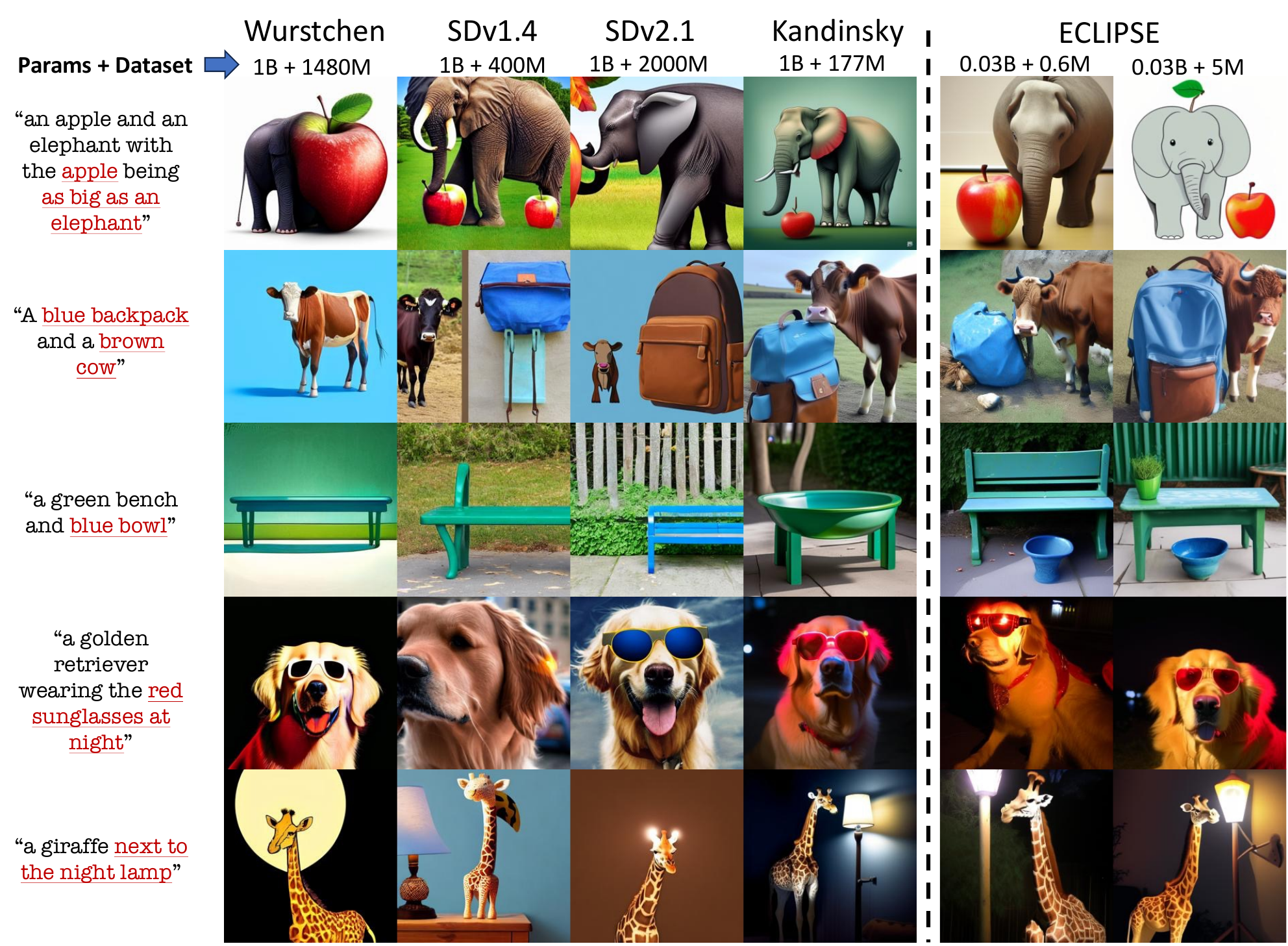}
    \caption{Qualitative result of our text-to-image prior, \eclipse, comparing with SOTA T2I model. Our prior model 
    reduces the model parameter requirements (from 1 Billion $\rightarrow$ 33 Million) and data requirements (from 177 Million $\rightarrow$ 5 Million $\rightarrow~$ 0.6 Million). Given this restrictive setting, \eclipse~ performs close to its huge counterpart (i.e., Kandinsky v2.2) and even outperforms models trained on huge datasets (i.e., Wurstchen, SDv1.4, and SDv2.1) in terms of compositions.
    }
    \label{fig:image_examples}
\end{figure*}

\medskip\noindent\textbf{Training and inference details.} 
We evaluate \eclipse~using two pre-trained image decoders: Karlo-v1-alpha and Kandinsky v2.2, trained on distinct CLIP vision encoders.
Our prior architecture is based on the standard PriorTransformer model~\cite{ramesh2022hierarchical}, modified to be time-independent.
The detailed architecture is outlined in the appendix. 
We configure prior models with 33 and 34 million parameters for Karlo and Kandinsky, respectively.
This contrasts with larger models in the field, which often use up to 1 billion parameters (as summarized in Table~\ref{tab:main_results}). 
The Projection, Diffusion-Baseline, and \eclipse~priors are trained for both diffusion image decoders, maintaining consistent hyperparameters (including total number of parameters) across all models.
Training on CC12M, CC3M, and LAION-HighRes is performed on 4 x RTX A6000 GPUs with a 256 per-GPU batch size, a learning rate of 0.00005, and the CosineAnnealingWarmRestarts scheduler~\cite{loshchilov2016sgdr}. 
Each model undergoes approximately 60,000 iterations, totaling around 200 GPU hours. 
For MSCOCO, training takes about 100 GPU hours. 
This can be further reduced to $\leq 50$ GPU hours if image-text pairs are preprocessed beforehand.
The diffusion prior is trained with a linear scheduler and 1000 DDPM timesteps. 
Inferences utilize 25 DDPM steps with 4.0 classifier-free guidance, while Projection and \eclipse~models do not require diffusion sampling. 
Image diffusion decoders are set to 50 DDIM steps and 7.5 classifier-free guidance.

\medskip\noindent\textbf{Evaluation setup.}
Our evaluation framework encompasses various metrics. We employ MS-COCO 30k to assess FID scores~\cite{heusel2017gans} and T2I-CompBench~\cite{huang2023t2i} for evaluating composition abilities in color, shape, texture, spatial, and non-spatial compositions. 
Given the impracticality of large-scale human studies, we approximate human preferences using PickScore~\cite{kirstain2023pick}, reporting results on the T2I-CompBench validation set comprising about 1500 unique prompts across different categories.

\begin{figure}[t]
    \centering

    % First subfigure
    \begin{subfigure}{\linewidth}
        \centering
        \includegraphics[width=\textwidth]{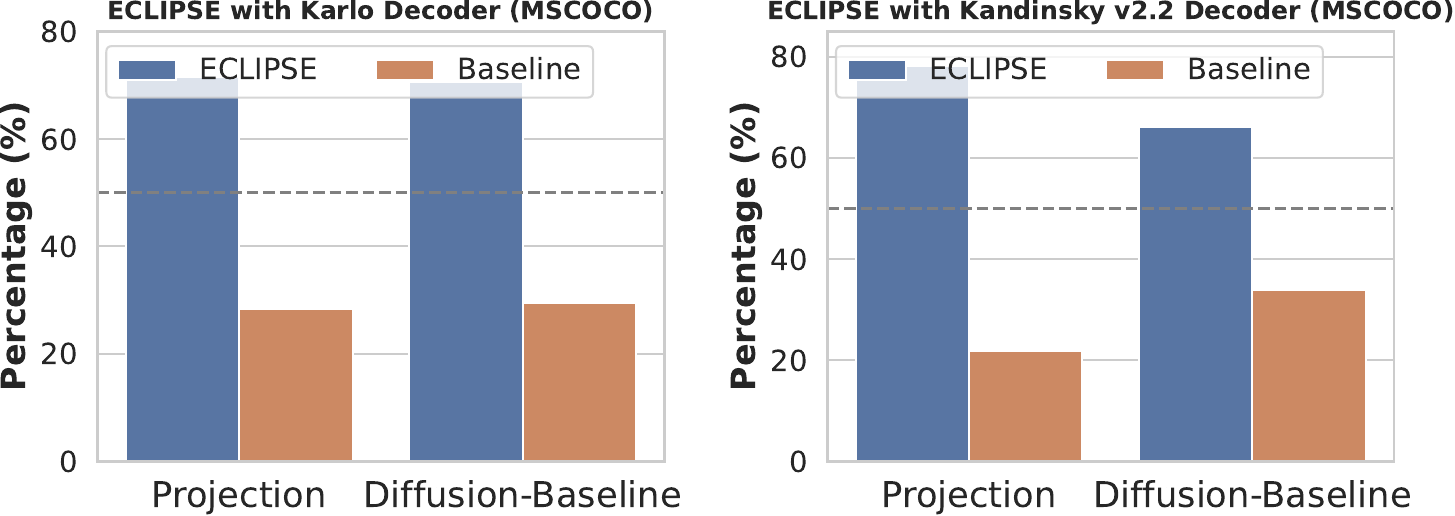}
        % \label{fig:sub1}
    \end{subfigure}
    % Second subfigure
    \begin{subfigure}{\linewidth}
        \centering
        \includegraphics[width=\linewidth]{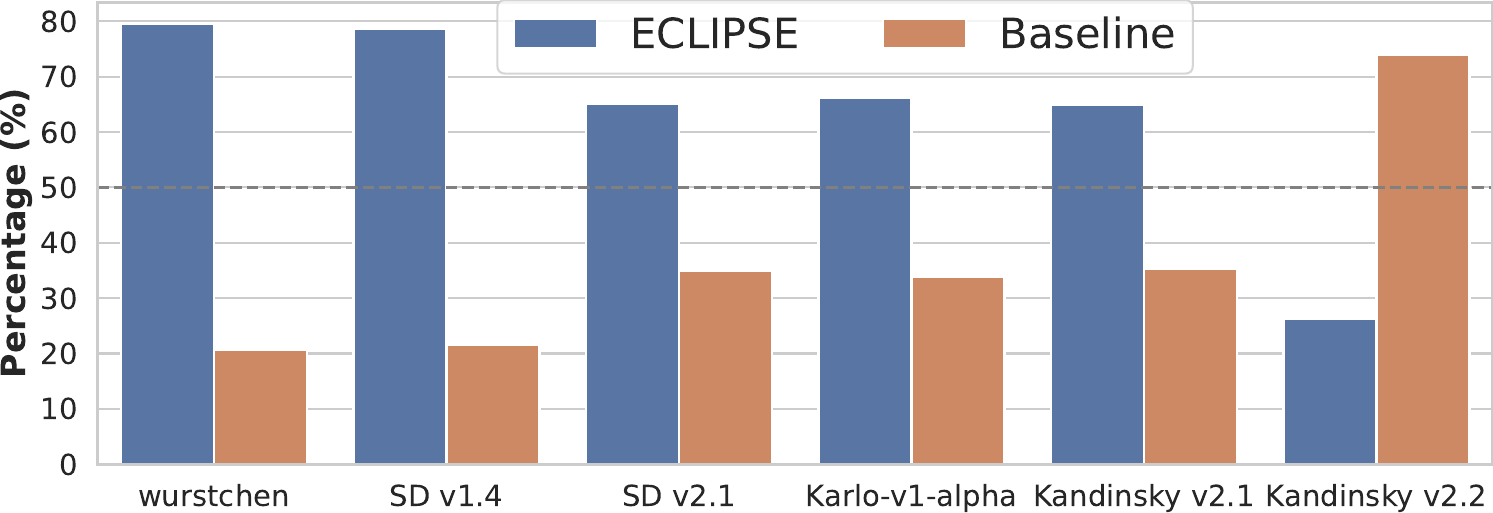}
        % \label{fig:sub2}
    \end{subfigure}
    
    \caption{
    Qualitative evaluations by human preferences approximated by the PickScore~\cite{kirstain2023pick}.
    The top two figures compare \eclipse~to Projection and Diffusion Baselines trained with the same amount of data and model size for both Karlo and Kandinsky decoders. In the bottom figure, we compare \eclipse~with the Kandinsky v2.2 decoder trained on the LAION-HighRes dataset against SOTA models.
    }
    \label{fig:qualitative_performance}
\end{figure}

\subsection{Quantitative Evaluations}
\label{sec:quantitative_evals}
In Table~\ref{tab:main_results}, we present a performance comparison between \eclipse~variants and leading T2I models. 
Our evaluation metrics include zero-shot Fréchet Inception Distance (FID) on MS-COCO 30k for image quality assessment and T2I-CompBench~\cite{huang2023t2i} for evaluating compositionality. 
\eclipse~priors, trained with both types of diffusion image decoders, demonstrate notable improvements.
\eclipse~consistently surpasses various baselines in terms of compositionality, irrespective of the dataset size. 
Its performance is comparable to that of DALL-E-2 and other SOTA models, a significant improvement considering \eclipse's parameter efficiency. 
Standard T2I priors usually incorporate 1 billion parameters, while \eclipse~operates with only 3.3\% of these parameters, maintaining competitive performance levels. 
When combined with corresponding diffusion image decoders, the total parameter count of \eclipse~is close to that of Stable Diffusion models, yet it outperforms them, especially considering that the latter are trained on a massive set of image-text pairs.
A noticeable decline in zero-shot FID (ZS-FID) is observed in comparison to the original Karlo. 
We attribute this variation to the image quality differences in the training dataset, suggesting a potential area for further investigation and improvement.
At the same time, if we utilize the smaller subset of high-resolution datasets then we can still maintain better FID and improve the compositions, as shown in the last row of Table~\ref{tab:main_results}.
\eclipse~prior with Kandinsky v2.2 decoder trained on LAION-HighRes subset achieves similar FID to other original Kandinsky v2.2 unCLIP model and at the same time outperforming in terms of compositions.

Table~\ref{tab:fine_grained_results} provides a comparison of various baseline training strategies for small prior models, using identical datasets and hyperparameters. 
\eclipse~exhibits superior performance across all datasets. 
We also note that diffusion priors benefit from larger datasets, supporting our premise that such priors necessitate extensive training data for optimal results, which is also attributed to the CFG. 
In contrast, \eclipse~demonstrates the consistent performance on compositions irrespective of the amount of image-text pairs.

\begin{table}[!t]
\centering
\caption{
Comparison of \eclipse~with respect to the various baseline prior learning strategies on four categories of composition prompts in the T2I-CompBench. All prior models are of 33 million parameters and trained on the same hyperparameters.
}

\label{tab:fine_grained_results}

\scriptsize

\resizebox{\linewidth}{!}{%

\begin{tabular}{l | cccc}\toprule

\multirow{2}{*}{\textbf{Methods}} &\multicolumn{4}{c}{\textbf{T2I-CompBench}} \\
 &\textbf{Color ($\uparrow$)} &\textbf{Shape ($\uparrow$)} &\textbf{Texture ($\uparrow$)} &\textbf{Spatial ($\uparrow$)}  \\\midrule

\textsuperscript{{\color{BrickRed}\textbf{MSCOCO}} with {\color{BrickRed}\textbf{Karlo}}} & & & \\[-3pt]

Projection & 0.4667 & 0.4421 & 0.5051 & 0.1478\\
Diffusion-Baseline & 0.4678 & 0.4797 & 0.4956 & 0.1240\\
\rowcolor{Apricot!50}\textbf{\eclipse~} &\textbf{0.5965} &\textbf{0.5063} &\textbf{0.6136} & \textbf{0.1574} \\

\midrule

\textsuperscript{{\color{BrickRed}\textbf{CC3M}} with {\color{BrickRed}\textbf{Karlo}}} & & & \\[-3pt]

Projection & 0.4362 & 0.4501 & 0.4948 & 0.1126\\
Diffusion-Baseline & \textbf{0.5493} & 0.4809 & 0.5462 & 0.1132\\
\rowcolor{Apricot!50}\textbf{\eclipse~} & 0.5421 & \textbf{0.5091} & \textbf{0.5881} & \textbf{0.1477} \\

\midrule

\textsuperscript{{\color{BrickRed}\textbf{CC12M}} with {\color{BrickRed}\textbf{Karlo}}} & & & \\[-3pt]

Projection  & 0.4659 & 0.4632 & 0.4995 & 0.1318\\
Diffusion-Baseline & 0.5390 & 0.4919 & 0.5276 & 0.1426\\
\rowcolor{Apricot!50}\textbf{\eclipse~} & \textbf{0.5660} & \textbf{0.5234} & \textbf{0.5941} & \textbf{0.1625}\\

\midrule
\midrule

\textsuperscript{{\color{BrickRed}\textbf{MSCOCO}} with {\color{BrickRed}\textbf{Kandinsky v2.2}}} & & & \\[-3pt]

Projection  & 0.4678 & 0.3736 & 0.4634 & 0.1268\\
Diffusion-Baseline  & 0.4646 & 0.4403 & 0.4834 & 0.1566\\
\rowcolor{Apricot!50}\textbf{\eclipse} & \textbf{0.5785} & \textbf{0.4951} & \textbf{0.6173} & \textbf{0.1794} \\

\midrule

\textsuperscript{{\color{BrickRed}\textbf{HighRes}} with {\color{BrickRed}\textbf{Kandinsky v2.2}}} &  \\[-3pt]

Projection & 0.5379 & 0.4983 & 0.5217 & 0.1573 \\
Diffusion-Baseline & 0.5706 &  0.5182 & 0.5067 & 0.1687 \\
\rowcolor{Apricot!50}\textbf{\eclipse} & \textbf{0.6119} & \textbf{0.5429} & \textbf{0.6165} & \textbf{0.1903} \\

\bottomrule

\end{tabular}

}

\end{table}

\subsection{Qualitative Evaluations}

In Figure~\ref{fig:image_examples}, we display qualitative examples from various methods responding to complex prompts. \eclipse~demonstrates superior performance in comparison to Stable Diffusion v1.4, Stable Diffusion v2.1, and Wurstchen, while closely matching the quality of its big counterpart, Kandinsky v2.2.
Interestingly, \eclipse~trained on only 0.6 million images maintains the compositions with minor degradation in image quality.
These observations align with our previously established quantitative results.
Beyond numerical metrics, understanding human preferences is crucial. 
To this end, we selected 1500 unique validation prompts from T2I-CompBench and assessed PickScore preferences. 
The results, illustrated in Figure~\ref{fig:qualitative_performance}, reveal that \eclipse~notably surpasses its baselines in respective restrictive settings with an average score of 71.6\%. 
We can also observe that the best \eclipse~variant (with Kandinsky decoder and trained on LAION-HighRes) consistently outperforms the other big SOTA models achieving an average performance of 63.36\%.
We observe that in terms of preferences, the original Kandinsky v2.2 diffusion prior (with a 1 billion parameter) trained on LAION-HighRes (175M) performs better than the \eclipse~prior (having 33 million parameters). 
We hypothesize that this might be due to its use of a large-scale dataset that contains more aesthetically pleasing images. 
We provide a set of qualitative results in the appendix to show that \eclipse~performs similarly well, if not better, \textit{w.r.t.} semantic understanding of the text.

\section{Analysis}
\label{sec:ablation}

\begin{figure}[t]
    \centering

    % First subfigure
    \begin{subfigure}{\linewidth}
        \centering
        \includegraphics[width=\textwidth]{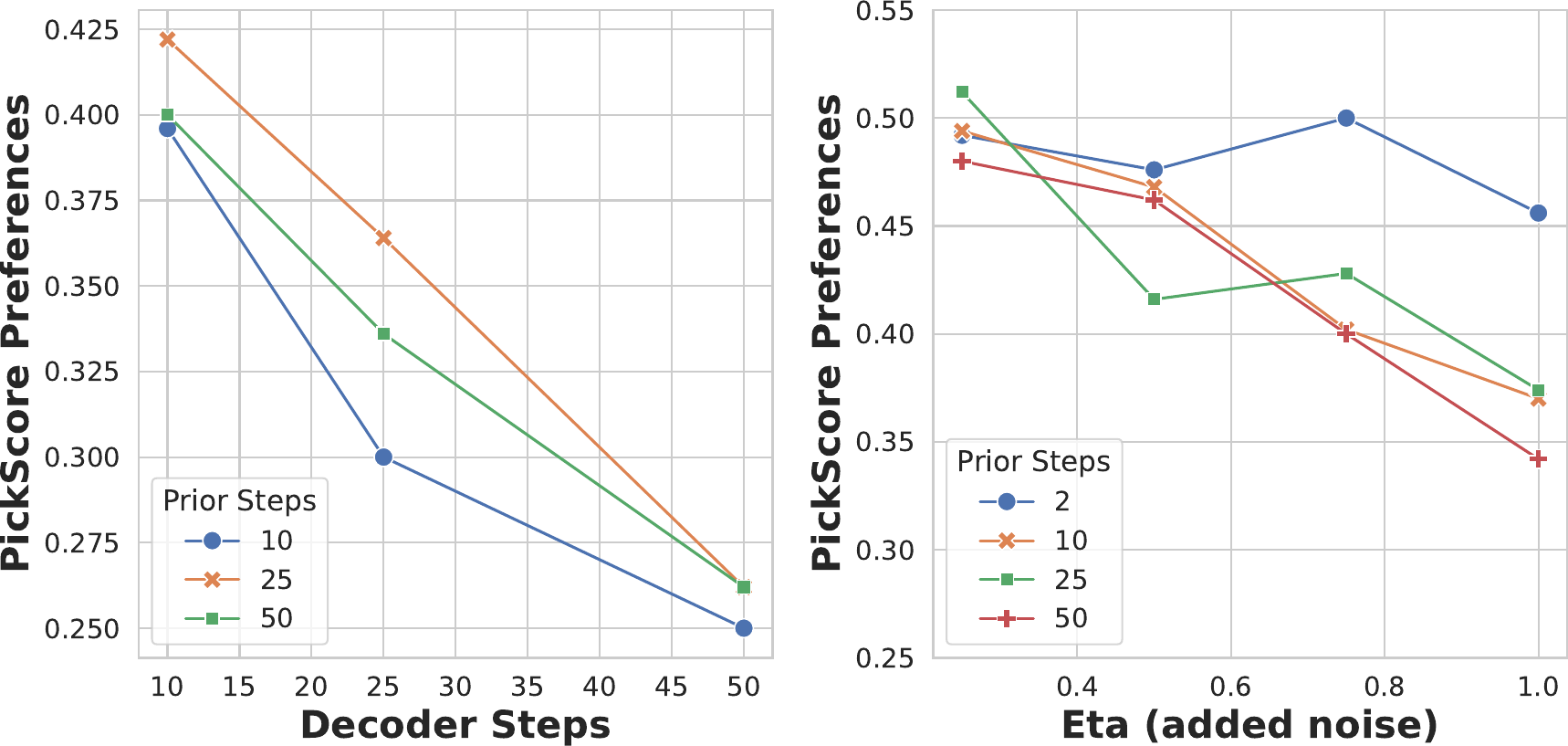}
        \caption{Left: Performance comparison by varying the prior steps and decoder steps \textit{w.r.t.} the fixed prior steps ($t=2$). Right: Performance comparison by varying the mean $\eta$ of the added scheduler noise ($\sigma_t\epsilon$)  \textit{w.r.t.} the noiseless predictions ($\eta=0$). Both experiments are on the Kandinsky v2.1.}
        % \label{fig:sub1}
    \end{subfigure}
    % Second subfigure
    \begin{subfigure}{\linewidth}
        \centering
        \includegraphics[width=\textwidth]{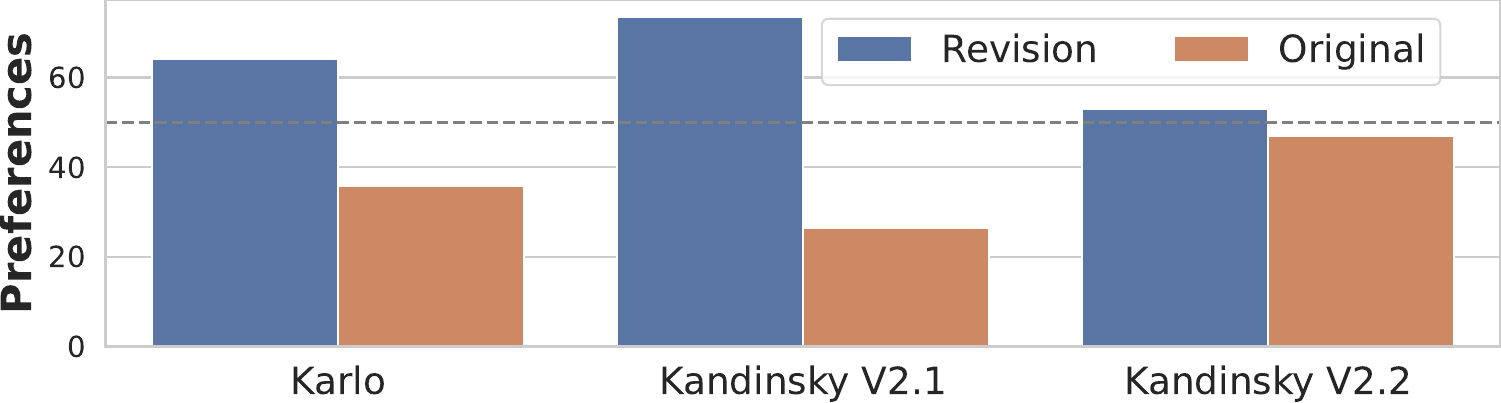}
        \caption{Overall performance comparisons on various pre-trained unCLIP models before and after reducing the prior steps to two and $\eta$ to $0.0$.}
        % \label{fig:sub2}
    \end{subfigure}
    
    \caption{Empirical analysis of the PickScore preferences of diffusion priors with respect to the various hyper-parameters.}
    \label{fig:prior_analysis}
\end{figure}

\noindent\textbf{Analyzing the traditional diffusion priors.}
To further support our choice of using non-diffusion prior models, we analyze the existing diffusion prior formulation. We conducted two key empirical studies:
1) Evaluating the Impact of Prior Steps: We examined how the number of prior steps influences model performance.
2) Assessing the Influence of Added Noise ($\sigma_t\epsilon$): We focused on understanding how the introduction of noise affects human preferences.
For these studies, we utilized PickScore preferences, and the outcomes, depicted in Figure~\ref{fig:prior_analysis}, corroborate our hypothesis: both the prior steps and the addition of ($\sigma_t\epsilon$) detrimentally affect performance.
Furthermore, as indicated in Table~\ref{tab:fine_grained_results}, diffusion prior surpasses the projection baseline if provided with more high-quality data. 
We attribute this enhanced performance to the incorporation of classifier-free guidance, which bolsters the model's generalization capabilities to a certain extent. 
However, it's worth noting that both baselines are still outperformed by \eclipse. 
This observation underscores the effectiveness of our proposed methodology in comparison to traditional approaches in the realm of T2I.

\begin{figure}[t]
    \centering

    % First subfigure
    \begin{subfigure}{\linewidth}
        \centering
        \includegraphics[width=\textwidth]{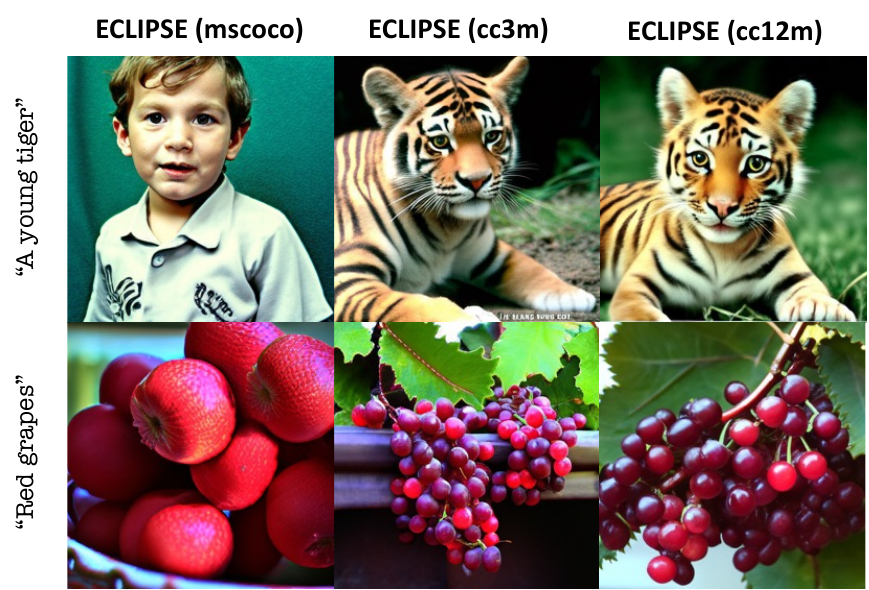}
        % \caption{First subfigure caption}
        % \label{fig:sub1}
    \end{subfigure}
    % Second subfigure
    \begin{subfigure}{\linewidth}
        \centering
        \includegraphics[width=\linewidth]{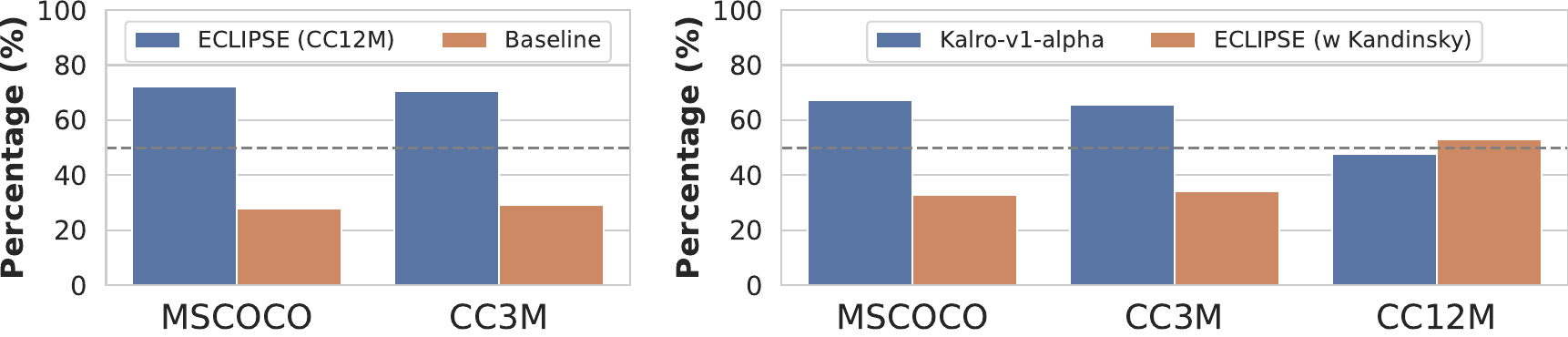}
        % \caption{Second subfigure caption.}
        % \label{fig:sub2}
    \end{subfigure}
    
    \caption{The top figure shows the qualitative examples of the biases learned by the T2I prior models.
    Bottom figures show the PickScore preferences of the \eclipse~models trained on various datasets with respect to the other datasets (left) and Karlo (right).}
    \label{fig:data_importance}
\end{figure}

\medskip\noindent\textbf{Importance of data selection.}
In our previous analysis (Table~\ref{tab:main_results} and~\ref{tab:fine_grained_results}), we demonstrated that \eclipse~attains competitive performance on composition benchmarks regardless of dataset size. 
This achievement is largely due to the integration of the contrastive loss $\mathcal{L}_{CLS}$ (Eq.\ref{eq:clip}). 
However, the final objective function also incorporates the $\mathcal{L}_{proj}$ (Eq.\ref{eq:projection}), which is pivotal in estimating the vision latent distribution.
This estimation is fundamentally dependent on the training distribution ($P_{XY}$), potentially leading the model to learn spurious correlations within $P_{XY}$. 
Consequently, the model's image quality could directly correlate with the overall quality of images in the training set.
To further substantiate this, we evaluated the preferences for \eclipse~models trained on MSCOCO, CC3M, and CC12M, in comparison to among themselves and Karlo-v1-alpha. 
The outcomes, presented in Figure~\ref{fig:data_importance}, reveal that the \eclipse~model trained on CC12M outperforms those trained on other datasets, exhibiting performance on par with big counterpart. 
\eclipse~prior (w Karlo decoder) trained on the CC12M dataset performs comparably to Karlo-v1-alpha while \eclipse~priors trained on other datasets struggle to do so; suggesting the importance of the high-quality data.
Furthermore, as illustrated in Figure~\ref{fig:data_importance}, the \eclipse~model trained on MSCOCO demonstrates a tendency to learn spurious correlations, such as associating the term ``young tiger'' with the person.

\section{Conclusion}

In this paper, we introduce a novel text-to-image prior learning strategy, named \eclipse, which leverages pre-trained vision-language models to provide additional supervision for training the prior model through contrastive learning. 
This approach significantly enhances the training efficiency of prior models in a parameter-efficient way.
Through comprehensive quantitative and qualitative evaluations, we assessed \eclipse~priors alongside various diffusion image decoders. 
The results indicate that \eclipse~surpasses both the baseline projection models and traditional diffusion-prior models.
Remarkably, \eclipse~achieves competitive performance alongside larger, state-of-the-art T2I models. 
It demonstrates that priors can be trained with merely 3.3\% of the parameters and 2.8\% of image-text pairs typically required, without compromising the performance. 
This advancement directly leads to at least 43\% overall compression of the unCLIP models.
Our findings show that pre-trained vision-language can be utilized more effectively; suggesting promising research direction where improving the vision-language models may directly benefit the T2I models.

\section*{Acknowledgement}
This work was supported by NSF RI grants \#1750082 and \#2132724, and a grant from Meta AI Learning Alliance. The views and opinions of the authors expressed herein do not necessarily state or reflect those of the funding agencies and employers.

{
    \small
    \bibliographystyle{ieeenat_fullname}
    \bibliography{main}

\begin{thebibliography}{48}
\providecommand{\natexlab}[1]{#1}
\providecommand{\url}[1]{\texttt{#1}}
\expandafter\ifx\csname urlstyle\endcsname\relax
  \providecommand{\doi}[1]{doi: #1}\else
  \providecommand{\doi}{doi: \begingroup \urlstyle{rm}\Url}\fi

\bibitem[Bakr et~al.(2023)Bakr, Sun, Shen, Khan, Li, and Elhoseiny]{bakr2023hrs}
Eslam~Mohamed Bakr, Pengzhan Sun, Xiaogian Shen, Faizan~Farooq Khan, Li~Erran Li, and Mohamed Elhoseiny.
\newblock Hrs-bench: Holistic, reliable and scalable benchmark for text-to-image models.
\newblock In \emph{Proceedings of the IEEE/CVF International Conference on Computer Vision}, pages 20041--20053, 2023.

\bibitem[Changpinyo et~al.(2021)Changpinyo, Sharma, Ding, and Soricut]{changpinyo2021conceptual}
Soravit Changpinyo, Piyush Sharma, Nan Ding, and Radu Soricut.
\newblock Conceptual 12m: Pushing web-scale image-text pre-training to recognize long-tail visual concepts.
\newblock In \emph{Proceedings of the IEEE/CVF Conference on Computer Vision and Pattern Recognition}, pages 3558--3568, 2021.

\bibitem[Chefer et~al.(2023)Chefer, Alaluf, Vinker, Wolf, and Cohen-Or]{chefer2023attend}
Hila Chefer, Yuval Alaluf, Yael Vinker, Lior Wolf, and Daniel Cohen-Or.
\newblock Attend-and-excite: Attention-based semantic guidance for text-to-image diffusion models.
\newblock \emph{ACM Transactions on Graphics (TOG)}, 42\penalty0 (4):\penalty0 1--10, 2023.

\bibitem[Chen et~al.(2023{\natexlab{a}})Chen, Yu, Ge, Yao, Xie, Wu, Wang, Kwok, Luo, Lu, et~al.]{chen2023pixart}
Junsong Chen, Jincheng Yu, Chongjian Ge, Lewei Yao, Enze Xie, Yue Wu, Zhongdao Wang, James Kwok, Ping Luo, Huchuan Lu, et~al.
\newblock Pixart-alpha: Fast training of diffusion transformer for photorealistic text-to-image synthesis.
\newblock \emph{arXiv preprint arXiv:2310.00426}, 2023{\natexlab{a}}.

\bibitem[Chen et~al.(2023{\natexlab{b}})Chen, Laina, and Vedaldi]{chen2023training}
Minghao Chen, Iro Laina, and Andrea Vedaldi.
\newblock Training-free layout control with cross-attention guidance.
\newblock \emph{arXiv preprint arXiv:2304.03373}, 2023{\natexlab{b}}.

\bibitem[Cong et~al.(2023)Cong, Min, Li, Rosenhahn, and Yang]{cong2023attribute}
Yuren Cong, Martin~Renqiang Min, Li~Erran Li, Bodo Rosenhahn, and Michael~Ying Yang.
\newblock Attribute-centric compositional text-to-image generation.
\newblock \emph{arXiv preprint arXiv:2301.01413}, 2023.

\bibitem[Donghoon et~al.(2022)Donghoon, Jiseob, Jisu, Jongmin, Minwoo, Woonhyuk, and Saehoon]{kakaobrain2022karlo-v1-alpha}
Lee Donghoon, Kim Jiseob, Choi Jisu, Kim Jongmin, Byeon Minwoo, Baek Woonhyuk, and Kim Saehoon.
\newblock Karlo-v1.0.alpha on coyo-100m and cc15m.
\newblock \url{https://github.com/kakaobrain/karlo}, 2022.

\bibitem[Fernandez et~al.(2023)Fernandez, Couairon, J{\'e}gou, Douze, and Furon]{stable_signature}
Pierre Fernandez, Guillaume Couairon, Herv{\'e} J{\'e}gou, Matthijs Douze, and Teddy Furon.
\newblock The stable signature: Rooting watermarks in latent diffusion models.
\newblock \emph{arXiv preprint arXiv:2303.15435}, 2023.

\bibitem[Gal et~al.(2022)Gal, Alaluf, Atzmon, Patashnik, Bermano, Chechik, and Cohen-or]{gal2022image}
Rinon Gal, Yuval Alaluf, Yuval Atzmon, Or Patashnik, Amit~Haim Bermano, Gal Chechik, and Daniel Cohen-or.
\newblock An image is worth one word: Personalizing text-to-image generation using textual inversion.
\newblock In \emph{The Eleventh International Conference on Learning Representations}, 2022.

\bibitem[Heusel et~al.(2017)Heusel, Ramsauer, Unterthiner, Nessler, and Hochreiter]{heusel2017gans}
Martin Heusel, Hubert Ramsauer, Thomas Unterthiner, Bernhard Nessler, and Sepp Hochreiter.
\newblock Gans trained by a two time-scale update rule converge to a local nash equilibrium.
\newblock \emph{Advances in neural information processing systems}, 30, 2017.

\bibitem[Ho and Salimans(2022)]{ho2022classifier}
Jonathan Ho and Tim Salimans.
\newblock Classifier-free diffusion guidance.
\newblock \emph{arXiv preprint arXiv:2207.12598}, 2022.

\bibitem[Ho et~al.(2020)Ho, Jain, and Abbeel]{ho2020denoising}
Jonathan Ho, Ajay Jain, and Pieter Abbeel.
\newblock Denoising diffusion probabilistic models.
\newblock \emph{Advances in neural information processing systems}, 33:\penalty0 6840--6851, 2020.

\bibitem[Huang et~al.(2023)Huang, Sun, Xie, Li, and Liu]{huang2023t2i}
Kaiyi Huang, Kaiyue Sun, Enze Xie, Zhenguo Li, and Xihui Liu.
\newblock T2i-compbench: A comprehensive benchmark for open-world compositional text-to-image generation.
\newblock \emph{arXiv preprint arXiv:2307.06350}, 2023.

\bibitem[Karras et~al.(2019)Karras, Laine, and Aila]{karras2019style}
Tero Karras, Samuli Laine, and Timo Aila.
\newblock A style-based generator architecture for generative adversarial networks.
\newblock In \emph{Proceedings of the IEEE/CVF conference on computer vision and pattern recognition}, pages 4401--4410, 2019.

\bibitem[Kim et~al.(2023{\natexlab{a}})Kim, Song, Castells, and Choi]{kim2023architectural}
Bo-Kyeong Kim, Hyoung-Kyu Song, Thibault Castells, and Shinkook Choi.
\newblock On architectural compression of text-to-image diffusion models.
\newblock \emph{arXiv preprint arXiv:2305.15798}, 2023{\natexlab{a}}.

\bibitem[Kim et~al.(2021)Kim, Ren, and Yang]{kim2020decentralized}
Changhoon Kim, Yi Ren, and Yezhou Yang.
\newblock Decentralized attribution of generative models.
\newblock In \emph{International Conference on Learning Representations}, 2021.

\bibitem[Kim et~al.(2023{\natexlab{b}})Kim, Min, Patel, Cheng, and Yang]{kim2023wouaf}
Changhoon Kim, Kyle Min, Maitreya Patel, Sheng Cheng, and Yezhou Yang.
\newblock Wouaf: Weight modulation for user attribution and fingerprinting in text-to-image diffusion models.
\newblock \emph{arXiv preprint arXiv:2306.04744}, 2023{\natexlab{b}}.

\bibitem[Kirstain et~al.(2023)Kirstain, Polyak, Singer, Matiana, Penna, and Levy]{kirstain2023pick}
Yuval Kirstain, Adam Polyak, Uriel Singer, Shahbuland Matiana, Joe Penna, and Omer Levy.
\newblock Pick-a-pic: An open dataset of user preferences for text-to-image generation.
\newblock \emph{arXiv preprint arXiv:2305.01569}, 2023.

\bibitem[Koh et~al.(2021)Koh, Baldridge, Lee, and Yang]{koh2021text}
Jing~Yu Koh, Jason Baldridge, Honglak Lee, and Yinfei Yang.
\newblock Text-to-image generation grounded by fine-grained user attention.
\newblock In \emph{Proceedings of the IEEE/CVF winter conference on applications of computer vision}, pages 237--246, 2021.

\bibitem[Li et~al.(2023{\natexlab{a}})Li, Liu, Wu, Mu, Yang, Gao, Li, and Lee]{li2023gligen}
Yuheng Li, Haotian Liu, Qingyang Wu, Fangzhou Mu, Jianwei Yang, Jianfeng Gao, Chunyuan Li, and Yong~Jae Lee.
\newblock Gligen: Open-set grounded text-to-image generation.
\newblock In \emph{Proceedings of the IEEE/CVF Conference on Computer Vision and Pattern Recognition}, pages 22511--22521, 2023{\natexlab{a}}.

\bibitem[Li et~al.(2023{\natexlab{b}})Li, Wang, Jin, Hu, Chemerys, Fu, Wang, Tulyakov, and Ren]{li2023snapfusion}
Yanyu Li, Huan Wang, Qing Jin, Ju Hu, Pavlo Chemerys, Yun Fu, Yanzhi Wang, Sergey Tulyakov, and Jian Ren.
\newblock Snapfusion: Text-to-image diffusion model on mobile devices within two seconds.
\newblock \emph{arXiv preprint arXiv:2306.00980}, 2023{\natexlab{b}}.

\bibitem[Li et~al.(2022)Li, Min, Li, and Xu]{li2022stylet2i}
Zhiheng Li, Martin~Renqiang Min, Kai Li, and Chenliang Xu.
\newblock Stylet2i: Toward compositional and high-fidelity text-to-image synthesis.
\newblock In \emph{Proceedings of the IEEE/CVF Conference on Computer Vision and Pattern Recognition}, pages 18197--18207, 2022.

\bibitem[Lin et~al.(2014)Lin, Maire, Belongie, Hays, Perona, Ramanan, Doll{\'a}r, and Zitnick]{lin2014microsoft}
Tsung-Yi Lin, Michael Maire, Serge Belongie, James Hays, Pietro Perona, Deva Ramanan, Piotr Doll{\'a}r, and C~Lawrence Zitnick.
\newblock Microsoft coco: Common objects in context.
\newblock In \emph{Computer Vision--ECCV 2014: 13th European Conference, Zurich, Switzerland, September 6-12, 2014, Proceedings, Part V 13}, pages 740--755. Springer, 2014.

\bibitem[Loshchilov and Hutter(2016)]{loshchilov2016sgdr}
Ilya Loshchilov and Frank Hutter.
\newblock Sgdr: Stochastic gradient descent with warm restarts.
\newblock In \emph{International Conference on Learning Representations}, 2016.

\bibitem[Luo et~al.(2023)Luo, Tan, Huang, Li, and Zhao]{luo2023latent}
Simian Luo, Yiqin Tan, Longbo Huang, Jian Li, and Hang Zhao.
\newblock Latent consistency models: Synthesizing high-resolution images with few-step inference.
\newblock \emph{arXiv preprint arXiv:2310.04378}, 2023.

\bibitem[Nichol et~al.(2021)Nichol, Dhariwal, Ramesh, Shyam, Mishkin, McGrew, Sutskever, and Chen]{nichol2021glide}
Alex Nichol, Prafulla Dhariwal, Aditya Ramesh, Pranav Shyam, Pamela Mishkin, Bob McGrew, Ilya Sutskever, and Mark Chen.
\newblock Glide: Towards photorealistic image generation and editing with text-guided diffusion models.
\newblock \emph{arXiv preprint arXiv:2112.10741}, 2021.

\bibitem[Nie et~al.(2023)Nie, Kim, Yang, and Ren]{nie2023attributing}
Guangyu Nie, Changhoon Kim, Yezhou Yang, and Yi Ren.
\newblock Attributing image generative models using latent fingerprints.
\newblock \emph{arXiv preprint arXiv:2304.09752}, 2023.

\bibitem[Okawa et~al.(2023)Okawa, Lubana, Dick, and Tanaka]{okawa2023compositional}
Maya Okawa, Ekdeep~Singh Lubana, Robert~P Dick, and Hidenori Tanaka.
\newblock Compositional abilities emerge multiplicatively: Exploring diffusion models on a synthetic task.
\newblock \emph{arXiv preprint arXiv:2310.09336}, 2023.

\bibitem[Patel et~al.(2023)Patel, Gokhale, Baral, and Yang]{patel2023conceptbed}
Maitreya Patel, Tejas Gokhale, Chitta Baral, and Yezhou Yang.
\newblock Conceptbed: Evaluating concept learning abilities of text-to-image diffusion models.
\newblock \emph{arXiv preprint arXiv:2306.04695}, 2023.

\bibitem[Peebles and Xie(2023)]{peebles2023scalable}
William Peebles and Saining Xie.
\newblock Scalable diffusion models with transformers.
\newblock In \emph{Proceedings of the IEEE/CVF International Conference on Computer Vision}, pages 4195--4205, 2023.

\bibitem[Pernias et~al.(2023)Pernias, Rampas, and Aubreville]{pernias2023wuerstchen}
Pablo Pernias, Dominic Rampas, and Marc Aubreville.
\newblock Wuerstchen: Efficient pretraining of text-to-image models.
\newblock \emph{arXiv preprint arXiv:2306.00637v2}, 2023.

\bibitem[Phung et~al.(2023)Phung, Ge, and Huang]{phung2023grounded}
Quynh Phung, Songwei Ge, and Jia-Bin Huang.
\newblock Grounded text-to-image synthesis with attention refocusing.
\newblock \emph{arXiv preprint arXiv:2306.05427}, 2023.

\bibitem[Radford et~al.(2021)Radford, Kim, Hallacy, Ramesh, Goh, Agarwal, Sastry, Askell, Mishkin, Clark, et~al.]{radford2021learning}
Alec Radford, Jong~Wook Kim, Chris Hallacy, Aditya Ramesh, Gabriel Goh, Sandhini Agarwal, Girish Sastry, Amanda Askell, Pamela Mishkin, Jack Clark, et~al.
\newblock Learning transferable visual models from natural language supervision.
\newblock In \emph{International conference on machine learning}, pages 8748--8763. PMLR, 2021.

\bibitem[Ramesh et~al.(2021)Ramesh, Pavlov, Goh, Gray, Voss, Radford, Chen, and Sutskever]{ramesh2021zero}
Aditya Ramesh, Mikhail Pavlov, Gabriel Goh, Scott Gray, Chelsea Voss, Alec Radford, Mark Chen, and Ilya Sutskever.
\newblock Zero-shot text-to-image generation.
\newblock In \emph{International Conference on Machine Learning}, pages 8821--8831. PMLR, 2021.

\bibitem[Ramesh et~al.(2022)Ramesh, Dhariwal, Nichol, Chu, and Chen]{ramesh2022hierarchical}
Aditya Ramesh, Prafulla Dhariwal, Alex Nichol, Casey Chu, and Mark Chen.
\newblock Hierarchical text-conditional image generation with clip latents.
\newblock \emph{arXiv preprint arXiv:2204.06125}, 1\penalty0 (2):\penalty0 3, 2022.

\bibitem[Razzhigaev et~al.(2023)Razzhigaev, Shakhmatov, Maltseva, Arkhipkin, Pavlov, Ryabov, Kuts, Panchenko, Kuznetsov, and Dimitrov]{razzhigaev2023kandinsky}
Anton Razzhigaev, Arseniy Shakhmatov, Anastasia Maltseva, Vladimir Arkhipkin, Igor Pavlov, Ilya Ryabov, Angelina Kuts, Alexander Panchenko, Andrey Kuznetsov, and Denis Dimitrov.
\newblock Kandinsky: an improved text-to-image synthesis with image prior and latent diffusion.
\newblock \emph{arXiv preprint arXiv:2310.03502}, 2023.

\bibitem[Rombach et~al.(2022)Rombach, Blattmann, Lorenz, Esser, and Ommer]{rombach2022high}
Robin Rombach, Andreas Blattmann, Dominik Lorenz, Patrick Esser, and Bj{\"o}rn Ommer.
\newblock High-resolution image synthesis with latent diffusion models.
\newblock In \emph{Proceedings of the IEEE/CVF conference on computer vision and pattern recognition}, pages 10684--10695, 2022.

\bibitem[Saharia et~al.(2022)Saharia, Chan, Saxena, Li, Whang, Denton, Ghasemipour, Gontijo~Lopes, Karagol~Ayan, Salimans, et~al.]{saharia2022photorealistic}
Chitwan Saharia, William Chan, Saurabh Saxena, Lala Li, Jay Whang, Emily~L Denton, Kamyar Ghasemipour, Raphael Gontijo~Lopes, Burcu Karagol~Ayan, Tim Salimans, et~al.
\newblock Photorealistic text-to-image diffusion models with deep language understanding.
\newblock \emph{Advances in Neural Information Processing Systems}, 35:\penalty0 36479--36494, 2022.

\bibitem[Salimans and Ho(2021)]{salimans2021progressive}
Tim Salimans and Jonathan Ho.
\newblock Progressive distillation for fast sampling of diffusion models.
\newblock In \emph{International Conference on Learning Representations}, 2021.

\bibitem[Schuhmann et~al.(2022)Schuhmann, Beaumont, Vencu, Gordon, Wightman, Cherti, Coombes, Katta, Mullis, Wortsman, et~al.]{schuhmann2022laion}
Christoph Schuhmann, Romain Beaumont, Richard Vencu, Cade Gordon, Ross Wightman, Mehdi Cherti, Theo Coombes, Aarush Katta, Clayton Mullis, Mitchell Wortsman, et~al.
\newblock Laion-5b: An open large-scale dataset for training next generation image-text models.
\newblock \emph{Advances in Neural Information Processing Systems}, 35:\penalty0 25278--25294, 2022.

\bibitem[Sharma et~al.(2018)Sharma, Ding, Goodman, and Soricut]{sharma2018conceptual}
Piyush Sharma, Nan Ding, Sebastian Goodman, and Radu Soricut.
\newblock Conceptual captions: A cleaned, hypernymed, image alt-text dataset for automatic image captioning.
\newblock In \emph{Proceedings of the 56th Annual Meeting of the Association for Computational Linguistics (Volume 1: Long Papers)}, pages 2556--2565, 2018.

\bibitem[Sohl-Dickstein et~al.(2015)Sohl-Dickstein, Weiss, Maheswaranathan, and Ganguli]{sohl2015deep}
Jascha Sohl-Dickstein, Eric Weiss, Niru Maheswaranathan, and Surya Ganguli.
\newblock Deep unsupervised learning using nonequilibrium thermodynamics.
\newblock In \emph{International conference on machine learning}, pages 2256--2265. PMLR, 2015.

\bibitem[Van Den~Oord et~al.(2017)Van Den~Oord, Vinyals, et~al.]{van2017neural}
Aaron Van Den~Oord, Oriol Vinyals, et~al.
\newblock Neural discrete representation learning.
\newblock \emph{Advances in neural information processing systems}, 30, 2017.

\bibitem[Vapnik(1991)]{vapnik1991principles}
Vladimir Vapnik.
\newblock Principles of risk minimization for learning theory.
\newblock \emph{Advances in neural information processing systems}, 4, 1991.

\bibitem[Zhai et~al.(2022)Zhai, Wang, Mustafa, Steiner, Keysers, Kolesnikov, and Beyer]{zhai2022lit}
Xiaohua Zhai, Xiao Wang, Basil Mustafa, Andreas Steiner, Daniel Keysers, Alexander Kolesnikov, and Lucas Beyer.
\newblock Lit: Zero-shot transfer with locked-image text tuning.
\newblock In \emph{Proceedings of the IEEE/CVF Conference on Computer Vision and Pattern Recognition}, pages 18123--18133, 2022.

\bibitem[Zhai et~al.(2023)Zhai, Mustafa, Kolesnikov, and Beyer]{zhai2023sigmoid}
Xiaohua Zhai, Basil Mustafa, Alexander Kolesnikov, and Lucas Beyer.
\newblock Sigmoid loss for language image pre-training.
\newblock \emph{arXiv preprint arXiv:2303.15343}, 2023.

\bibitem[Zhang et~al.(2017)Zhang, Xu, Li, Zhang, Wang, Huang, and Metaxas]{zhang2017stackgan}
Han Zhang, Tao Xu, Hongsheng Li, Shaoting Zhang, Xiaogang Wang, Xiaolei Huang, and Dimitris~N Metaxas.
\newblock Stackgan: Text to photo-realistic image synthesis with stacked generative adversarial networks.
\newblock In \emph{Proceedings of the IEEE international conference on computer vision}, pages 5907--5915, 2017.

\bibitem[Zhou et~al.(2022)Zhou, Zhang, Chen, Li, Tensmeyer, Yu, Gu, Xu, and Sun]{zhou2022towards}
Yufan Zhou, Ruiyi Zhang, Changyou Chen, Chunyuan Li, Chris Tensmeyer, Tong Yu, Jiuxiang Gu, Jinhui Xu, and Tong Sun.
\newblock Towards language-free training for text-to-image generation.
\newblock In \emph{Proceedings of the IEEE/CVF Conference on Computer Vision and Pattern Recognition}, pages 17907--17917, 2022.

\end{thebibliography}
}

\appendix
\clearpage
\maketitlesupplementary

\section{Implementation Details}
\label{sec:implementation}

Table~\ref{tab:architecture} shows the comparison between \eclipse, Karlo, and Kaninsky priors. 
Notably, \eclipse~prior uses very compressed architecture across the possible avenues (i.e., number of layers, number of attention heads, attention head dimension, etc.).
Karlo uses CLIP-Vit-L/14 with 768 projection dimensions.
While Kandinsky v2.2 uses the ViT-bigG-14-laion2B-39B-b160k with 1280 projection dimensions.
Overall, the total number of parameters in \eclipse~priors is about 33 million compared to 1 billion parameters of Karlo/Kandinsky priors.
Additionally, Projection and Diffusion-Baseline use the same architecture as \eclipse~prior for better comparisons. 
Except the Diffusion-Prior contains the additional time embeddings for diffusion modeling.

\begin{table}[!h]
\centering
\resizebox{\linewidth}{!}{%
\begin{tabular}{l| cc}
& \multirow{2}{*}{\textbf{\eclipse}}  & \textbf{Karlo / Kandinsky} \\
& & \textbf{Priors} \\
                      \midrule
Num Attention Heads   & 16       & 32                      \\
Attention Head Dim    & 32       & 64                      \\
Num Layers            & 10       & 20                      \\
Embedding Dim         & 768/1280 & 768/1280                \\
Additional Embeddings & 3        & 4                       \\
Dropout               & 0.0      & 0.0                     \\
Time Embed            & No       & Yes                     \\
\midrule
Total Parameters      & 33/34 M  & 1 B \\       \bottomrule

\end{tabular}%
}
\caption{
Prior model architecture hyperparameter details.
}
\label{tab:architecture}
\end{table}

\section{Hyper-parameter Analysis}
\label{sec:hyperparameters}

\begin{figure}
    \centering
    \includegraphics[width=\linewidth]{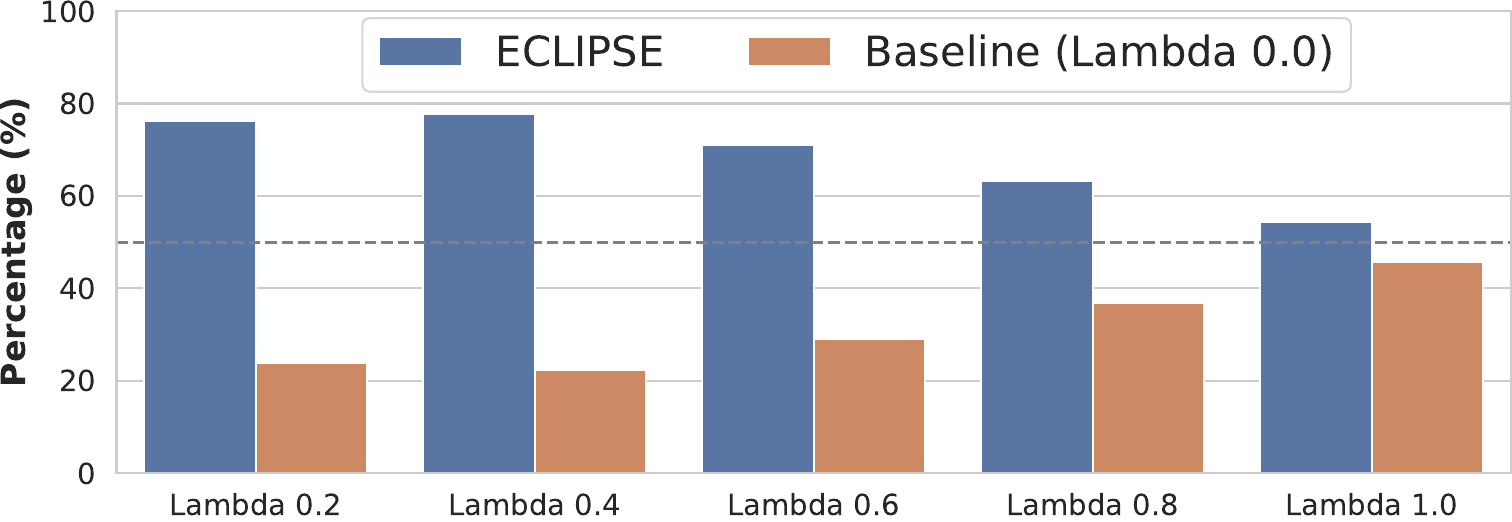}
    \caption{Hyperparameter ($\lambda$) ablation. This figure illustrates the PickScore preferences across the \eclipse~priors trained with different values of $\lambda$ \textit{w.r.t.} the Projection baseline (with $\lambda = 0.0$).}
    \label{fig:pickscore_hyperpara}
\end{figure}

\begin{figure}
    \centering
    \includegraphics[width=\linewidth]{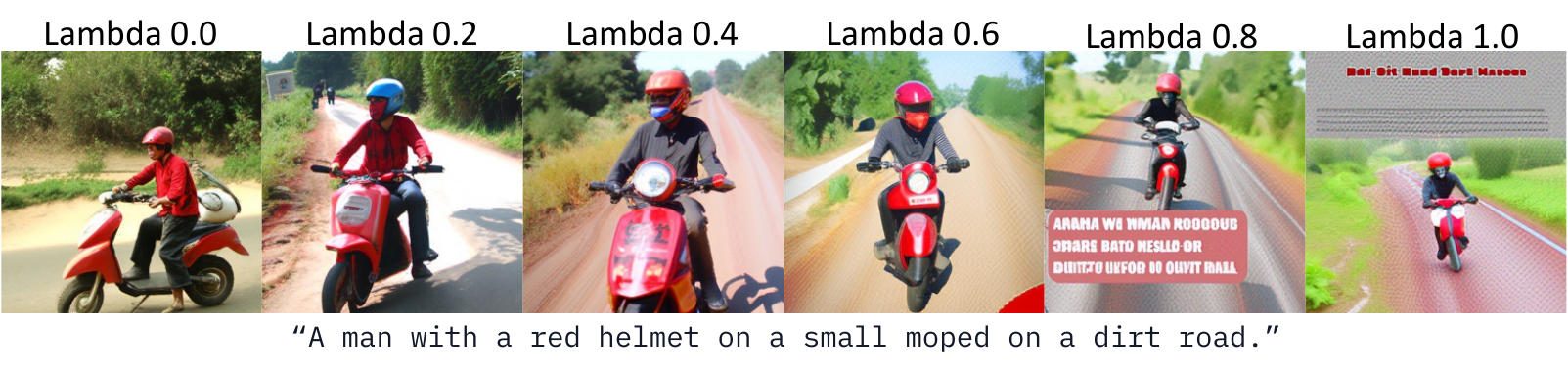}
    \caption{Qualitative example for \eclipse~priors (with Karlo decoder) trained with different values of hyperparameter ($\lambda$).}
    \label{fig:hyperparameter_qualitative}
\end{figure}

\eclipse~only contains one important hyperparameter ($\lambda$) that controls the contrastive learning. 
As discussed in Section 3.3, a higher value of $\lambda$ can make the prior model learn the different distribution that is highly aligned with text distributions. 
A lower value of $\lambda$ may not benefit in terms of generalization to unseen prompts. 
Hence, we conducted a small study on the MSCOCO dataset.
We train the \eclipse~priors for Karlo decoder on 20,000 iterations with the OneCycle learning rate. 
Figure~\ref{fig:pickscore_hyperpara} illustrates the pickscore preferences on T2I-CompBench of various values of $\lambda$. 
It can be observed that higher values of $\lambda$ lead to the same performance as the baseline. 
While lower values of $\lambda$ outperform the baseline by significant margins.
Additionally, Figure~\ref{fig:hyperparameter_qualitative} shows one qualitative example across the range of $\lambda$.
It can be seen that the generated image quality drops as $\lambda$ increases. 
\textbf{Hence, the optimal range is: $\lambda \in [0.2, 0.4]$.}

\section{\textbf{\eclipse~}Prior Model Scaling Behaviour}
\label{sec:scaling}

To analyze the scaling behavior of different prior learning strategies to a certain extent, we increase the prior model size from 33M to 89M. 
Table~\ref{tab:model_scale_results} shows the results when small and large priors are trained on the same dataset (CC12M) with the Karlo image diffusion decoder.
We train both versions of the prior models on 60,000 iterations (about 350 GPU hours) with exactly the same hyperparameters.
First, we observe that \textbf{\eclipse~prior improves the performance slightly}. 
Second, the Projection baseline gets the same performance, which suggests that \textbf{data is the bottleneck for the Projection prior}.
Third, interestingly Diffusion prior degrades the performance.
Upon further inspection, we found that 60,000 iterations are insufficient for the Diffusion model to converge.
Therefore, this verifies that \textbf{Diffusion-priors are resource-hungry}.
Importantly, \eclipse~priors easily converge irrespective of the data and number of parameters; suggesting that \eclipse~do not depend upon the huge resource constraints.

\begin{table}[!t]
\centering
\caption{
This table illustrates the scaling behavior of various T2I prior learning strategies. ``Small'' priors are 33 million in terms of parameters. And ``Large'' priors have 89 million parameters.
All prior models are trained on the CC12M dataset with the Karlo diffusion image decoder.
}

\label{tab:model_scale_results}

\scriptsize

\resizebox{\linewidth}{!}{%

\begin{tabular}{l | c | cccc}\toprule

\multirow{2}{*}{\textbf{Methods}} & \textbf{ZS} &\multicolumn{4}{c}{\textbf{T2I-CompBench}} \\
 & \textbf{FID} &\textbf{Color ($\uparrow$)} &\textbf{Shape ($\uparrow$)} &\textbf{Texture ($\uparrow$)} &\textbf{Spatial ($\uparrow$)}  \\\midrule

\textsuperscript{{\color{BrickRed}\textbf{33M Priors}}} & & & \\[-3pt]

Projection & 28.84 & 0.4659 & 0.4632 & 0.4995 & 0.1318\\
Diffusion-Baseline & \textbf{26.13} & 0.5390 & 0.4919 & 0.5276 & 0.1426\\
\rowcolor{Apricot!50}\textbf{\eclipse~} & 26.98 & \textbf{0.5660} & \textbf{0.5234} & \textbf{0.5941} & \textbf{0.1625}\\

\midrule

\textsuperscript{{\color{BrickRed}\textbf{89M Priors}}} & & & \\[-3pt]

Projection & 28.81 & 0.4579 & 0.4625 & 0.4761 & 0.1343\\
Diffusion-Baseline & 29.78 & 0.4988 & 0.4790 & 0.4604 & 0.1247\\
\rowcolor{Apricot!50}\textbf{\eclipse~} & \textbf{25.77} & \textbf{0.5712} & \textbf{0.5358} & \textbf{0.6194} & \textbf{0.16665}\\

\bottomrule

\end{tabular}

}

\end{table}

\section{Aesthetics: Kandinsky v2.2 \textit{vs.} \eclipse}
\label{sec:aesthetics}

As was observed in Figure~\ref{fig:qualitative_performance} from the main paper, the Kandinsky v2.2 model outperforms the \eclipse~prior when evaluated in terms of human preferences measured by Pickscore. 
We attribute this behavior to the differences in the aesthetic quality of the generated images. 
Therefore, we conduct additional actual human studies to analyze this behavior further.
In total, we randomly selected 200 prompts from the MSCOCO validation set (instead of T2I-CompBench as reported in Figure~\ref{fig:qualitative_performance}) and asked the human evaluators to perform two studies:

\begin{figure}[h]
    \centering
    \includegraphics[width=\linewidth]{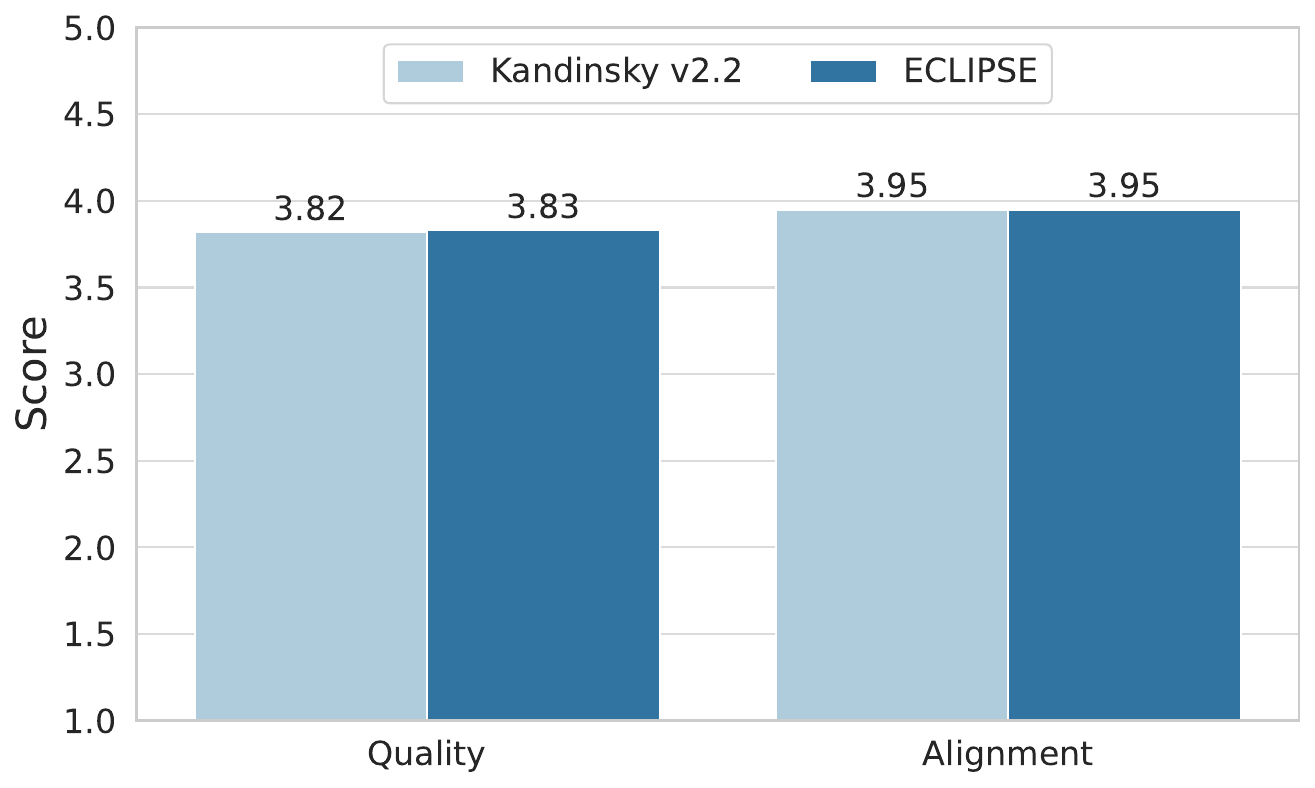}
    \caption{Human evaluations of the \eclipse~\textit{vs.}Kandinsky v2.2 generated images. It can be observed that both models are rated equally in terms of image quality and caption alignment.}
    \label{fig:human_eval_score}
\end{figure}

\begin{figure}[h]
    \centering
    \includegraphics[width=\linewidth]{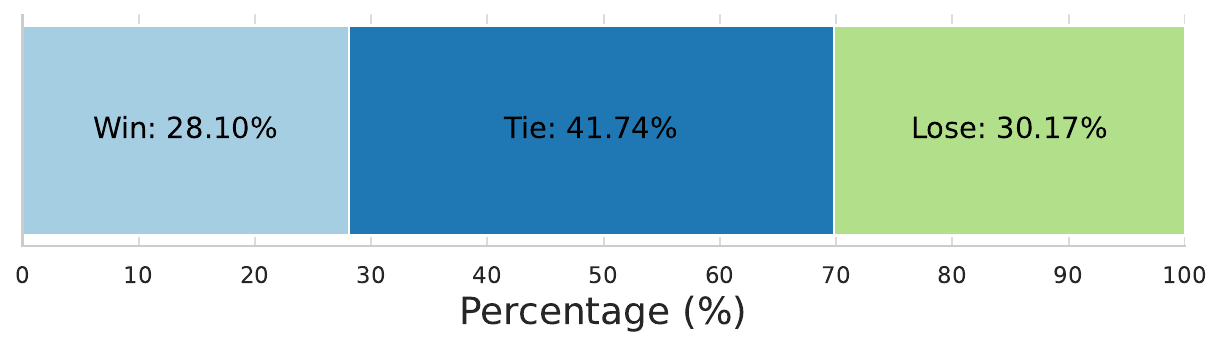}
    \caption{This figure illustrates the human preferences between \eclipse~prior for Kandinsky model (trained on LAION-HighRes subset) \textit{vs.} Original Kandinsky v2.2 model.}
    \label{fig:perference_human_eval}
\end{figure}

\begin{itemize}
    \item Rate each image in terms of quality and caption alignment between 1-5. Where 1 is the artificial-looking image and caption alignment is poor. While 5 represents a very high-quality image and is perfectly aligned with the captions.
    \item Image preferences in terms of aesthetics. We show images from both models and ask the evaluators to choose one which looks more aesthetically pleasing.
\end{itemize}

\begin{figure}[!t]
    \centering
    \includegraphics[width=\linewidth]{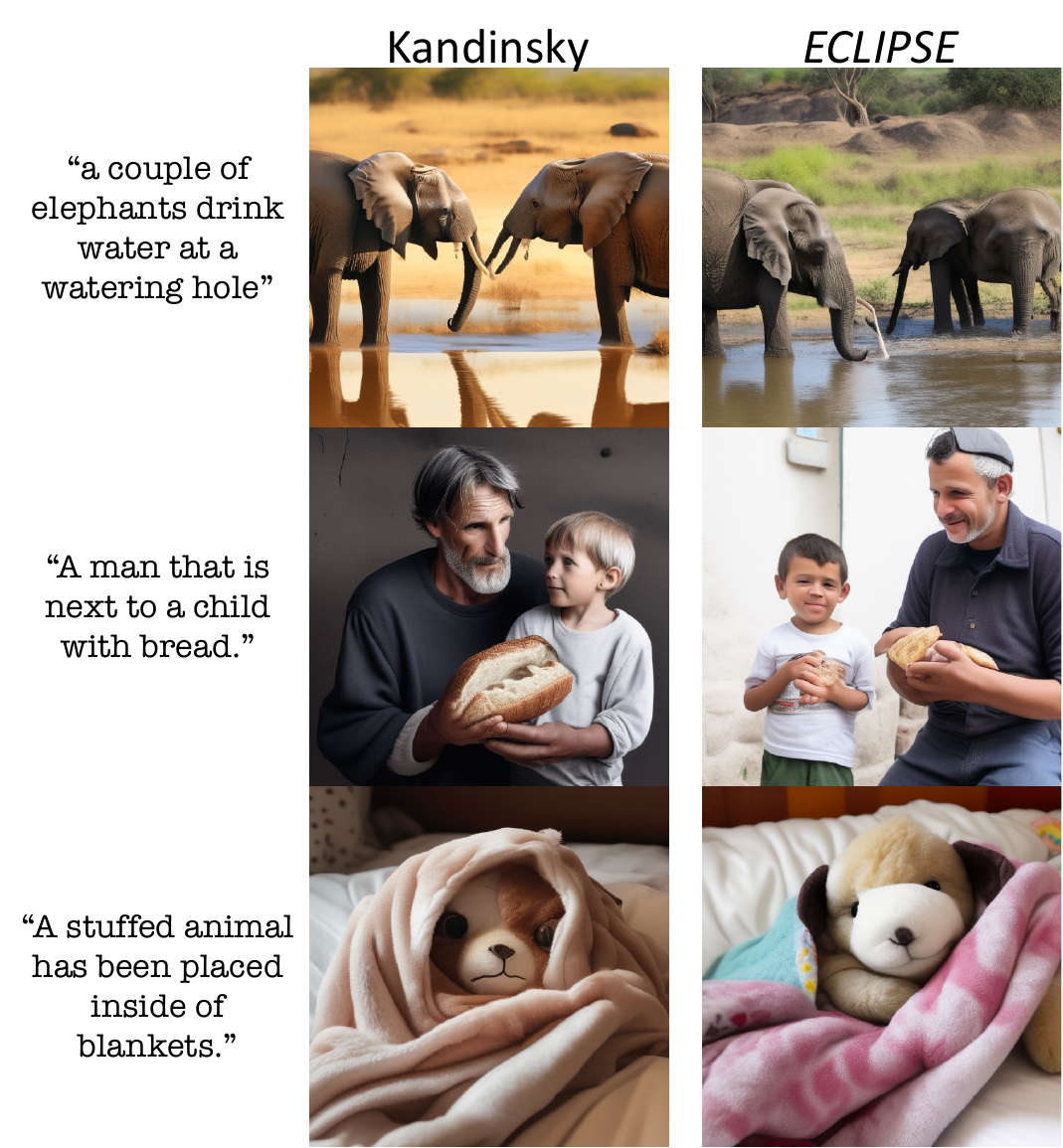}
    \caption{Qualitative examples comparing (in terms of aesthetics) \eclipse~with Kandinsky v2.2.}
    \label{fig:aesthetic_qualitative}
\end{figure}

\begin{figure*}
    \centering
    \includegraphics[width=\linewidth]{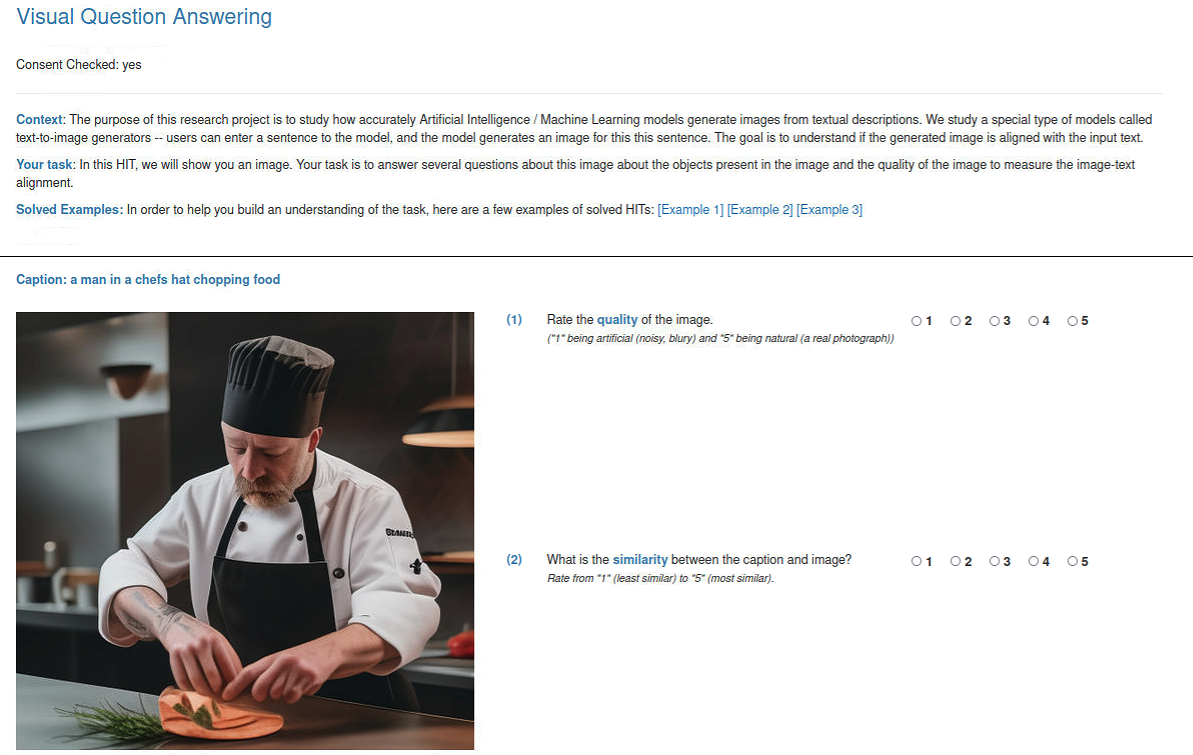}
    \caption{An example of human annotation for determining the image quality and caption alignment.}
    \label{fig:humaneval_score}
\end{figure*}

\begin{figure*}
    \centering
    \includegraphics[width=\linewidth]{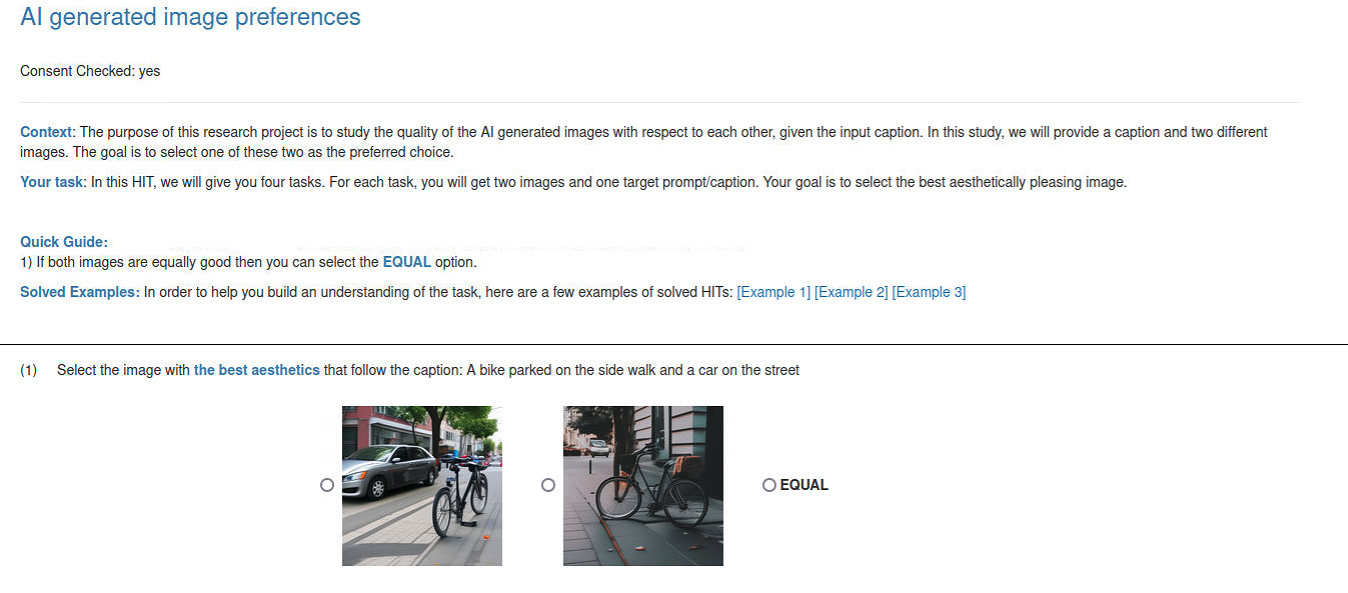}
    \caption{An example of human annotation for determining the most aesthetic image.}
    \label{fig:humaneval_preferences}
\end{figure*}

Interestingly, as shown in Figure~\ref{fig:human_eval_score}, both models are rated equally when evaluated independently. 
Additionally, according to Figure~\ref{fig:perference_human_eval}, Kandinsky v2.2 is preferred slightly more than the \eclipse~in terms of aesthetic quality.
This finding suggests that smaller prior trained with \eclipse~can perform equally (if not better) to those big prior models.
Figure~\ref{fig:aesthetic_qualitative} shares three examples from the MSCOCO. 
Both models perform equally well but Kandinsky is more aesthetically pleasing. 
Figure~\ref{fig:humaneval_score} and~\ref{fig:humaneval_preferences} show the MTurk human evaluation instructions.

\section{Diversity With Non-Diffusion Priors}
\label{sec:diversity}
\begin{figure}
    \centering
    \includegraphics[width=\linewidth]{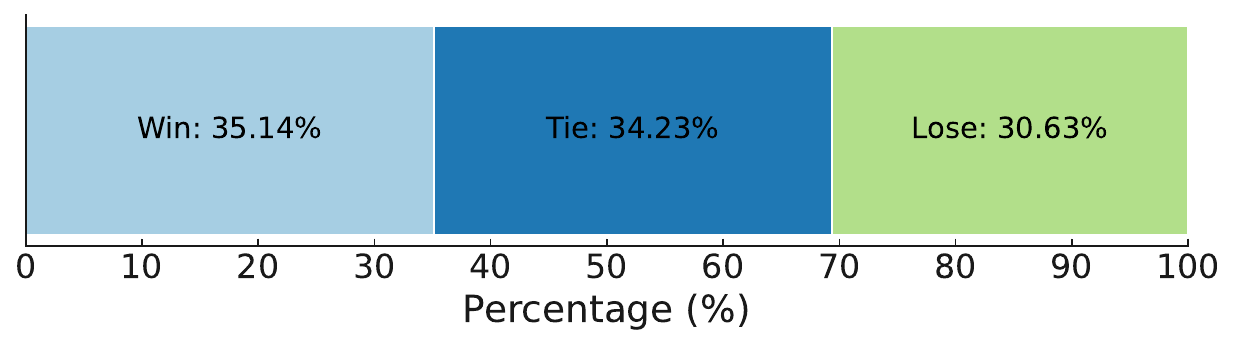}
    \caption{This figure illustrates the human preferences on the diversity of generated images between \eclipse~prior with Kandinsky v2.2 diffusion image decoder \textit{vs.} Kandinsky v2.2.}
    \label{fig:diversity_preferences}
\end{figure}

One important aspect of the diffusion models is the diversity of the generated images. 
Therefore, diversity and caption alignment go hand-in-hand. 
We further analyze whether having the non-diffusion prior hurts diversity or not. 
We perform additional qualitative evaluations and given a prompt -- we ask the human evaluators to select which of the two grids of six images are more diverse. 
This experiment is performed between \eclipse~and Kandinsky v2.2.
As shown in Figure~\ref{fig:diversity_preferences}, even if we use the non-diffusion prior model it does not hurt the diversity.
Diffusion image decoder is the main reason that contributes to the diversity and having diffusion or non-diffusion prior does not contribute that significantly. 

\section{More Qualitative Evaluations}
\label{sec:qualitative}

In this section, we provide more qualitative examples and discuss them. 
We also provide comparisons based on the diffusion image decoder used (i.e., Karlo and Kandinsky v2.2).
Finally, we discuss several failure cases.

\subsection{\textbf{\eclipse}~with Karlo Decoder}

\begin{figure*}
    \centering
    \includegraphics[width=\textwidth]{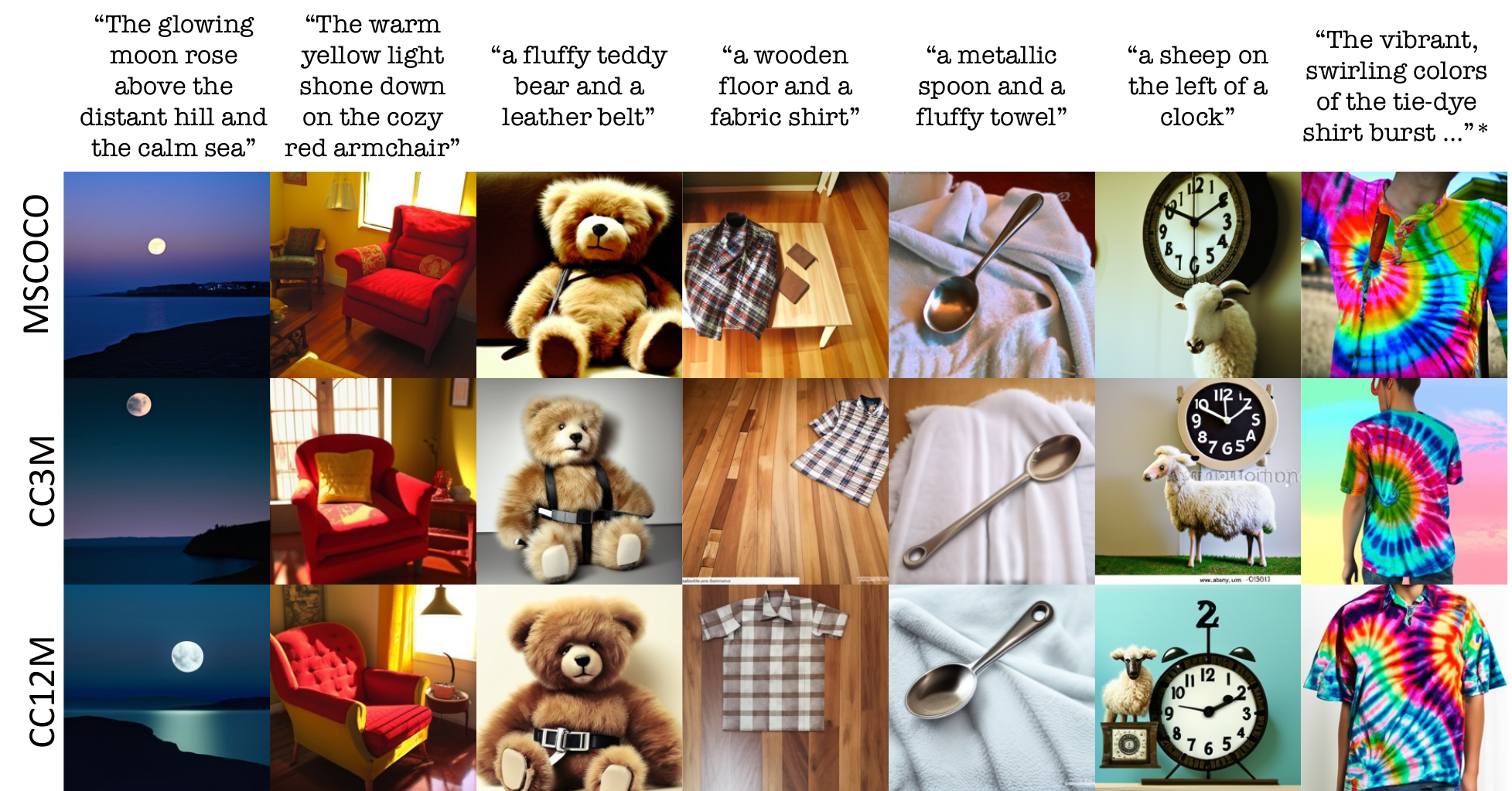}
    \caption{Qualitative comparisons of \eclipse~priors with Karlo decoder trained on different datasets. \textsuperscript{*} prompt is: "The vibrant, swirling colors of the tie-dye shirt burst with energy and personality, a unique expression of individuality and creativity."}
    \label{fig:karlo_datasets_qualitative}
\end{figure*}

\begin{figure*}
    \centering
    \includegraphics[width=\textwidth]{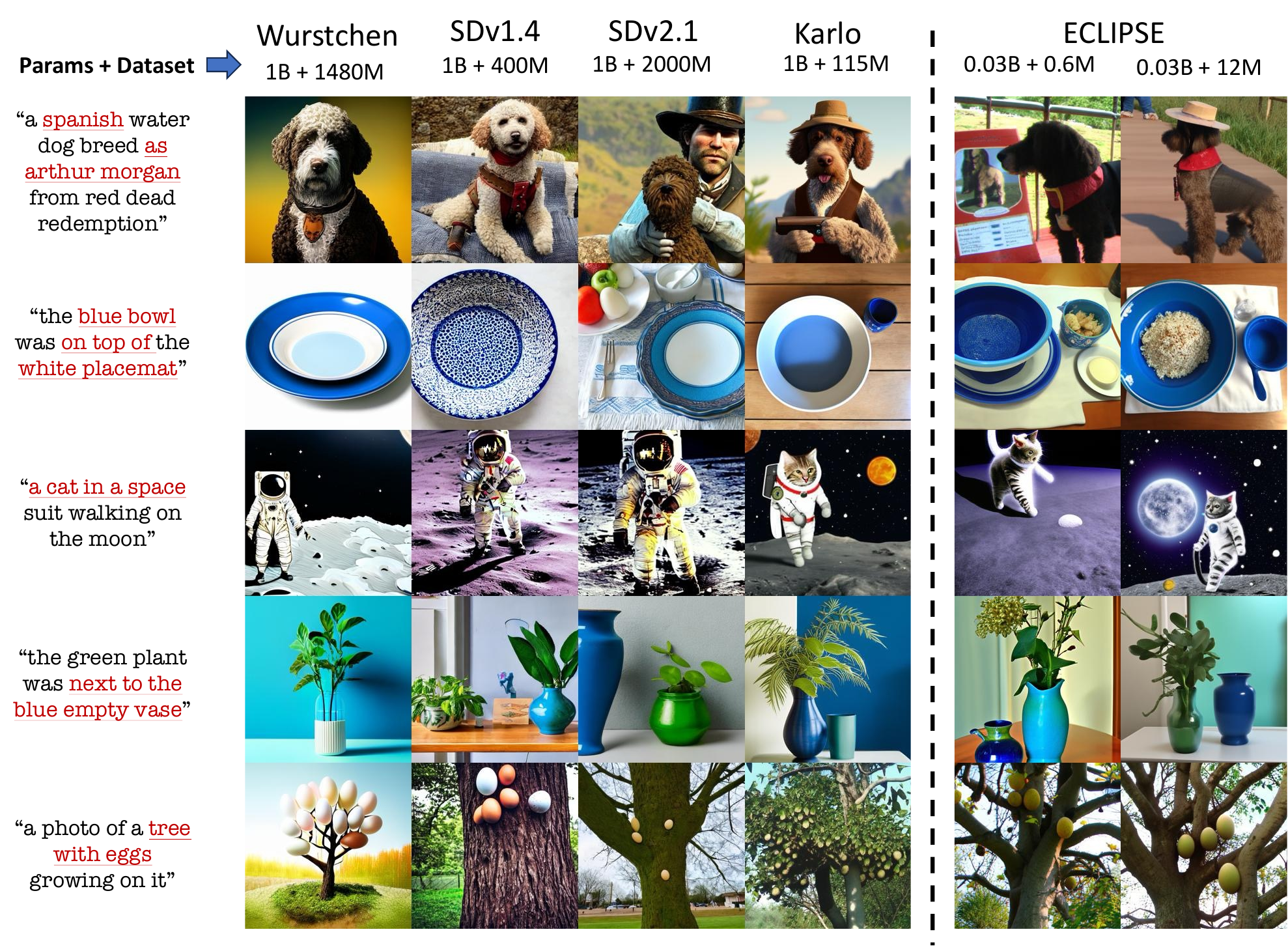}
    \caption{Qualitative result of our text-to-image prior, \eclipse~(with Karlo decoder), along with a comparison with SOTA T2I models. Our prior model reduces the prior
parameter requirements (from 1 Billion → 33 Million) and data requirements (from 115 Million → 12 Million → 0.6 Million).}
    \label{fig:extra_karlo_qualitative}
\end{figure*}

Figure~\ref{fig:karlo_baselines_qualitative} illustrates the comparison between Projection, Diffusion-Baseline, and \eclipse~priors trained on CC12M.
It can be seen that \eclipse~performs very well on complex composition prompts. 
While Projection and Diffusion baselines struggle to generate images aligned with the target prompt.
Figure~\ref{fig:karlo_datasets_qualitative} compares the \eclipse~priors trained on different datasets.
Here, \eclipse~prior trained on MSCOCO does not always follow the target prompt accurately and generates the lower quality images.
That said, the overall performance between all priors is very similar; suggesting that even a small amount of dataset is sufficient to distill the knowledge from the pre-trained Vision-Language models.
Figure~\ref{fig:extra_karlo_qualitative} compares the \eclipse~models with various SOTA methods.
Noticeably, \eclipse~performs better than the other baselines in terms of the ability to follow the target prompts.
For instance, many SOTA models cannot generate ``empty blue vase'', ``cat in space suit'', and ``blue bowl on white placemat''. 
Although we observe that \eclipse~prior trained with MSCOCO does follow the target text prompts but cannot generate high-quality images, which aligns with our previous findings.

\subsection{\textbf{\eclipse}~with Kandinsky Decoder}

\begin{figure*}
    \centering
    \includegraphics[width=\textwidth]{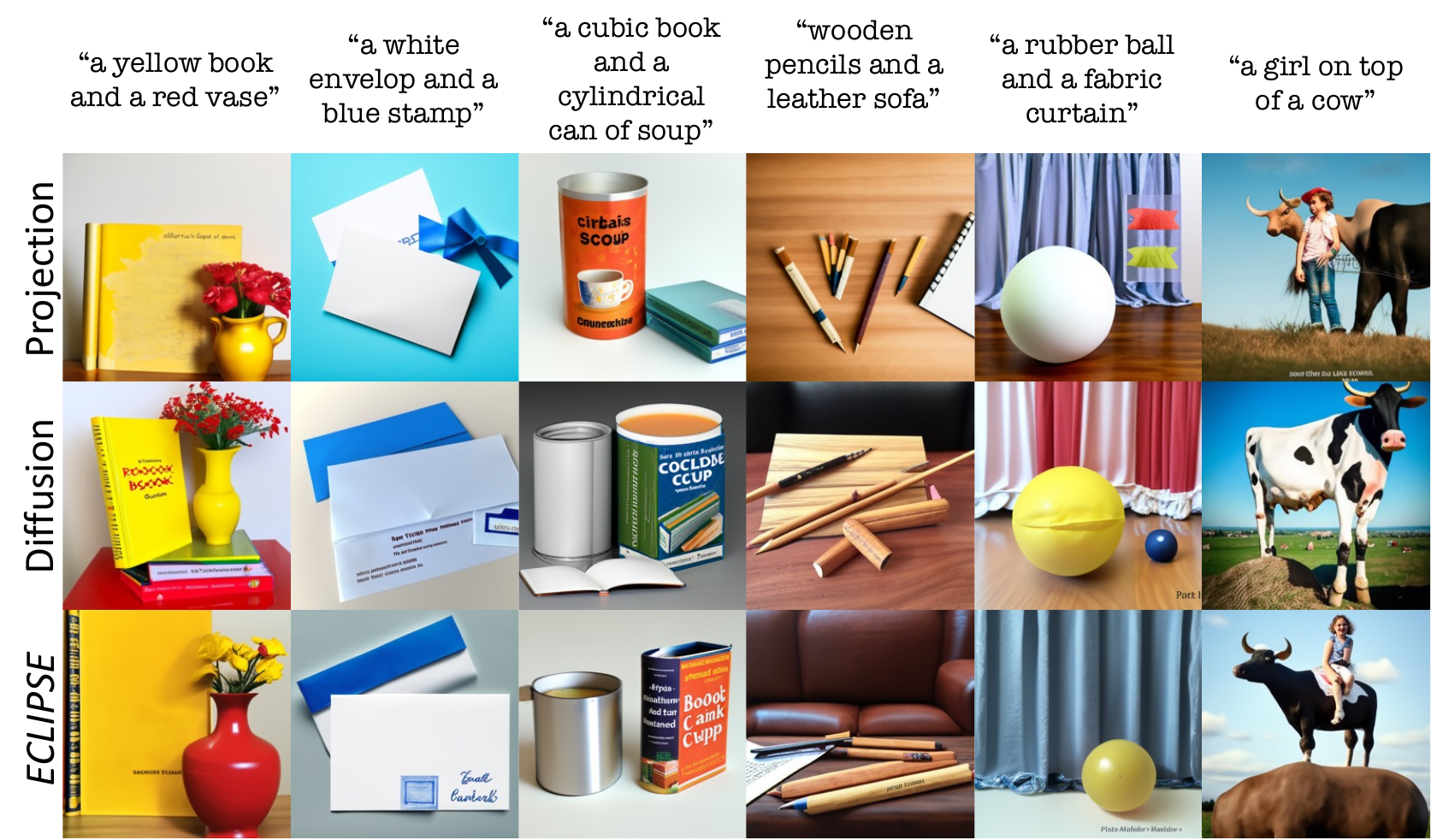}
    \caption{Qualitative comparisons between \eclipse~and baseline priors (having 33 million parameters) trained on  CC12M dataset with Karlo decoder. \textsuperscript{*} prompt is: "The bold, striking contrast of the black and white photograph captured the sense of the moment, a timeless treasure memory."}
    \label{fig:karlo_baselines_qualitative}
\end{figure*}

\begin{figure*}
    \centering
    \includegraphics[width=\textwidth]{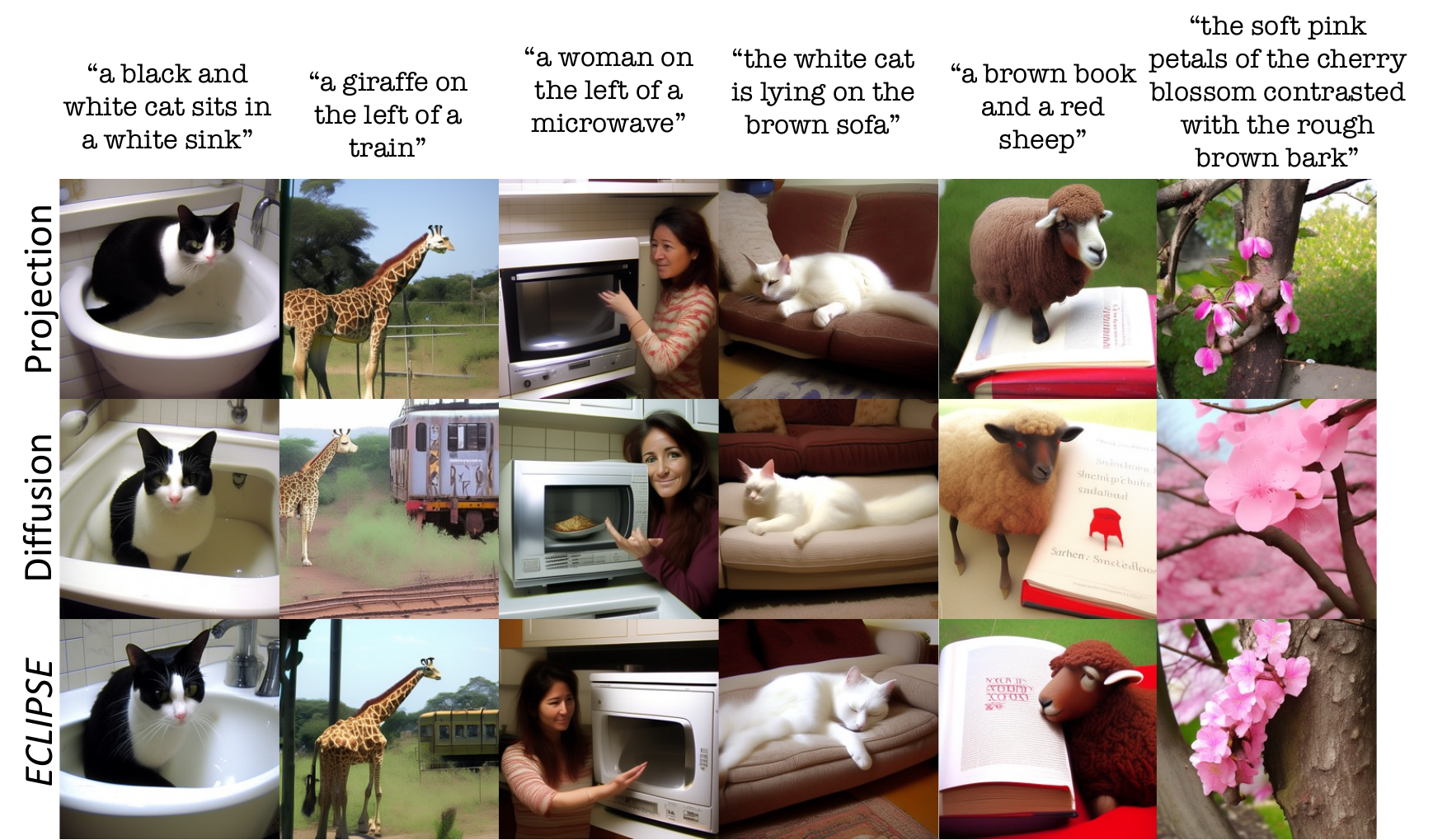}
    \caption{Qualitative comparisons between \eclipse~and baseline priors (having 34 million parameters) trained on  LAION-HighRes subset dataset with Kandinsky v2.2 diffusion image decoder.}
    \label{fig:kandinsky_baselines_qualitative}
\end{figure*}

Similarly, we analyze the qualitative results on Kandinsky diffusion image decoders.
Figure~\ref{fig:kandinsky_baselines_qualitative} compares the various baselines priors with the \eclipse~prior. 
We observe that baselines perform very closely to the \eclipse~prior, which is the opposite of what we found in Figure~\ref{fig:karlo_baselines_qualitative}.
We attribute this behavior to the change in the pre-trained CLIP encoder. 
Additionally, as shown in Table 2 of the main paper, both baseline priors perform very highly compared to the same priors trained on the CC12M dataset for the Karlo decoder. 
The only difference is the pre-trained vision-language model.
\textbf{Therefore, the selection of the Vision-Language model also plays a crucial role.}

\begin{figure*}
    \centering
    \includegraphics[width=\textwidth]{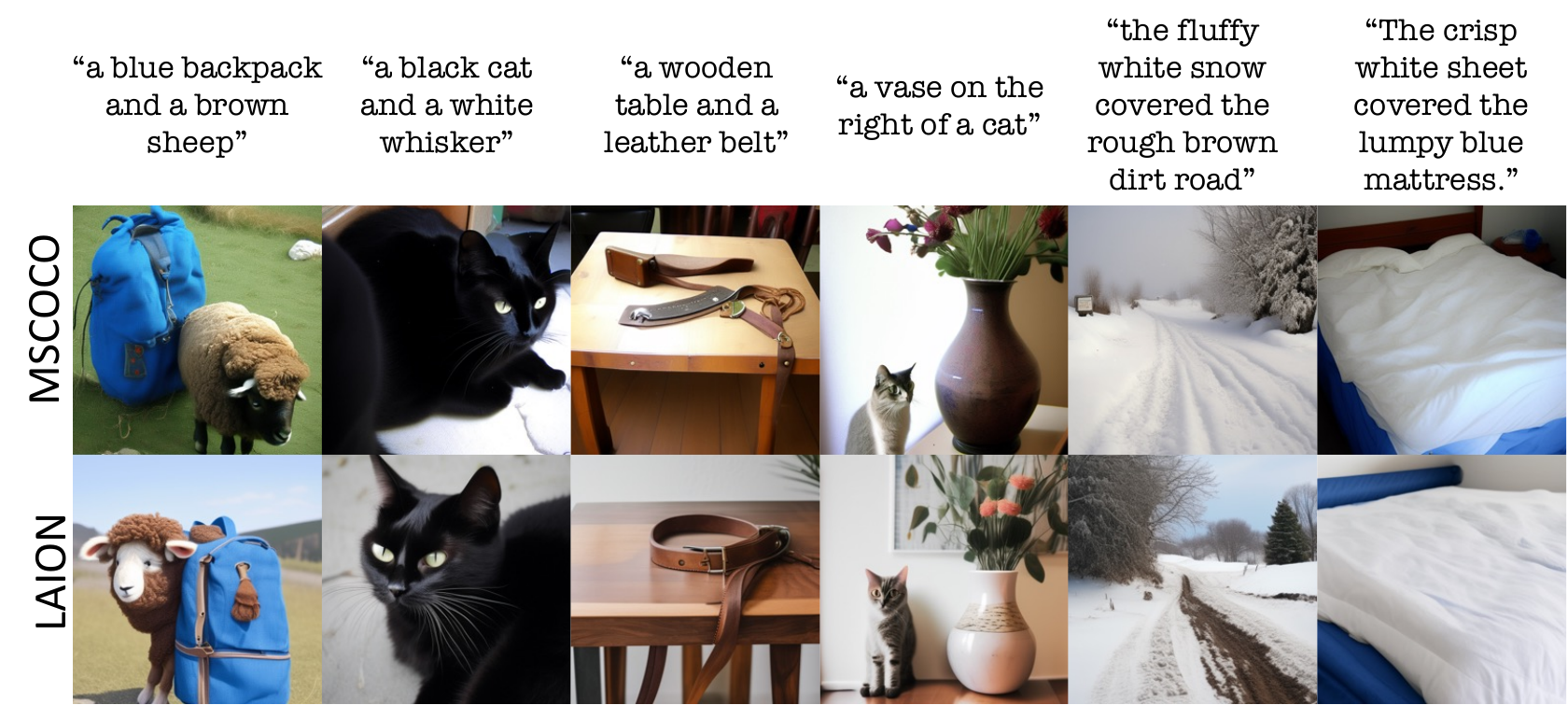}
    \caption{Qualitative comparisons between \eclipse~prior trained on MSCOCO and LAION datasets with Kandinsky v2.2 decoder.}
    \label{fig:kandinsky_datasets_qualitative}
\end{figure*}

\begin{figure*}
    \centering
    \includegraphics[width=\textwidth]{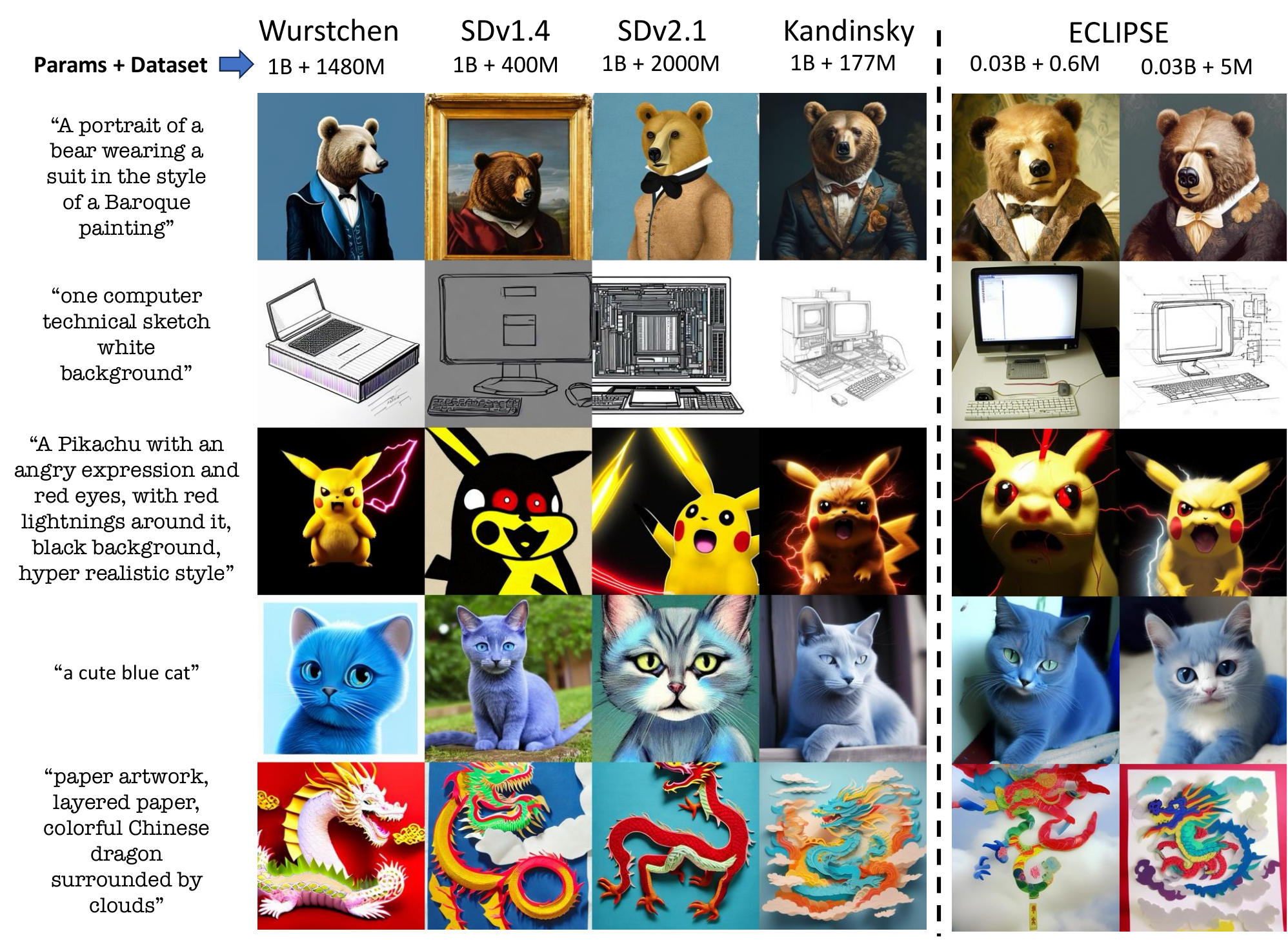}
    \caption{More qualitative result of our text-to-image prior, \eclipse~(with Kandinsky v2.2 decoder), along with a comparison with SOTA T2I models. Our prior model reduces the prior
parameter requirements (from 1 Billion → 33 Million) and data requirements (from 177 Million → 5 Million → 0.6 Million).}
    \label{fig:extra_kandinsky_qualitative}
\end{figure*}

Figure~\ref{fig:kandinsky_datasets_qualitative} illustrates the comparison with \eclipse~priors trained with different datasets.
It can be observed that with the use of the LAION-HighRes dataset not only did image quality improve but small intrinsic details (such as ``backpack'', ``belt'', etc.) also improved.
Even in some cases, prior training on the LAION subset performs better as the increase in the amount of data improves the performance.
Figure~\ref{fig:extra_kandinsky_qualitative} provides more qualitative examples to compare the \eclipse~priors with other respective SOTA methods.
As also previously observed, \eclipse~prior trained on LAION subset performs very close to the Kandinsky v2.2 in terms of following the text prompts.
While big SOTA models such as Stable Diffusion v1.4/2.1, and Wurstchen fall short despite being trained on millions of data.

\subsection{Failure Cases}

\begin{figure*}
    \centering
    \includegraphics[width=\textwidth]{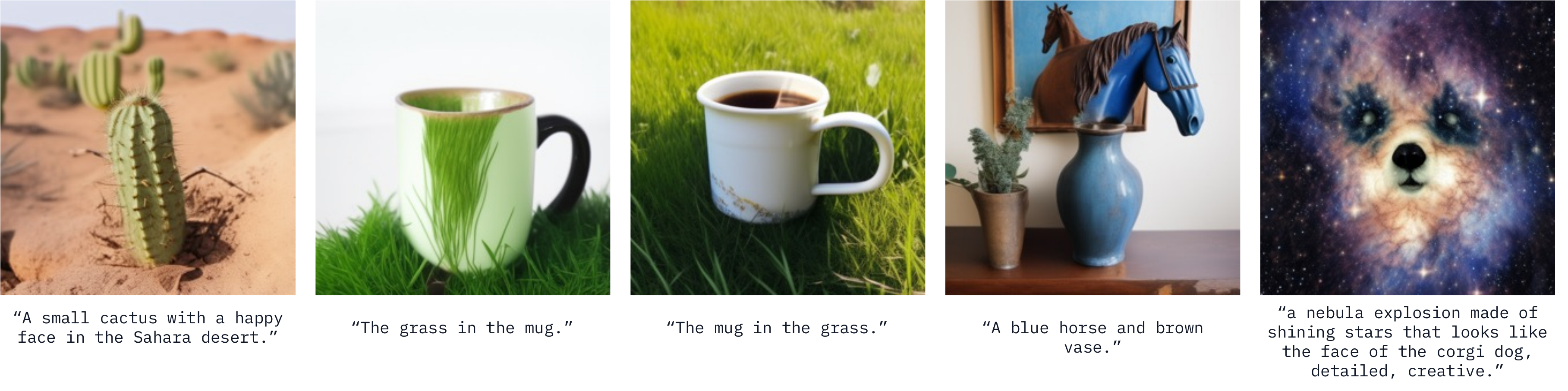}
    \caption{Instances where \eclipse~encounters the challenges in following the target text prompts.}
    \label{fig:challenges}
\end{figure*}

Figure~\ref{fig:challenges} shows some examples where \eclipse~model fails to follow the prompt precisely.
It is still difficult for the prior to learn something very unconventional. 
The model fails at generating some composition prompts (first four images).
It has been shown that vision-language models also suffer from such composition understanding (e.g., ``grass in the mug'' \textit{vs.} ``mug in the grass'').
Therefore, improving the Vision-Language model can further improve the capabilities of unCLIP priors.
Notably, \eclipse~finds it difficult to generate artistic imaginary images (such as ``nebula explosion that looks like corgi'').
However, such corner cases can be only solved with more diverse high-quality datasets.

\section{Future Work}
\label{sec:future}

In this work, we focus on improving text-to-image priors.
We assume that there exists a pre-trained diffusion image decoder that can be used as it is. 
To further improve the parameter efficiency for training, several relevant works on knowledge distillation and model compression can help.
Moreover, to improve the compositional abilities for unCLIP models, a better vision-language model (such as SigLIP) as the base model can be utilized to train the prior model using \eclipse. 
However, this will require the diffusion image decoder to be adjusted according to the new vision latent space.
We leave this direction as the future work as our paper primarily focuses on enhancing T2I priors.

\end{document}